\documentclass[11pt,a4paper]{article}
\usepackage[T1]{fontenc}
\usepackage[utf8]{inputenc}
\usepackage{lmodern}
\usepackage{microtype}
\usepackage{geometry}
\usepackage{setspace}
\usepackage{graphicx}
\usepackage{booktabs}
\usepackage{longtable}
\usepackage{array}
\usepackage{lscape}
\usepackage{enumitem}
\usepackage{xcolor}
\usepackage{amsmath,amssymb}
\usepackage{caption}
\usepackage[numbers,sort&compress]{natbib}
\usepackage{xurl}
\usepackage{hyperref}
\usepackage[nameinlink,noabbrev]{cleveref}

\definecolor{IndexBlue}{HTML}{1F5AA6}
\newcommand{\suppfiguretocshort}[2]{\expandafter\gdef\csname suppfiguretoc@#1\endcsname{#2}}
\newcommand{\suppfiguretoctitle}[2]{\ifcsname suppfiguretoc@#1\endcsname\csname suppfiguretoc@#1\endcsname\else#2\fi}
\newcommand{\contentscontinued}{\noindent{\large\bfseries Contents (continued)}\par\vspace{1em}}
\newcommand{\suppfigurepage}[4]{%
  \clearpage
  \newgeometry{left=0.975cm,right=0.975cm,top=0.5cm,bottom=1.2cm,footskip=0.7cm}%
  \thispagestyle{plain}%
  \phantomsection
  \addcontentsline{toc}{subsection}{Fig. #1. \suppfiguretoctitle{#1}{#3}}%
  \hypertarget{fig:#1}{}%
  \noindent\includegraphics{#2}\par
  \vspace{3mm}%
  \noindent{\fontsize{10}{12}\selectfont\textbf{Supplementary Fig. #1 | #3.} #4\par}%
  \vfill
  \clearpage
  \restoregeometry
}

\makeatletter
\renewcommand*\l@subsection{\@dottedtocline{2}{1.5em}{0em}}
\makeatother
\setlist{nosep,leftmargin=*}
\renewcommand{\bibsection}{\section{Supplementary References}}

\usepackage{url}
\DeclareCaptionLabelSeparator{bar}{\space\textbar\space}
\newsavebox{\MainCaptionBox}
\newlength{\MainGraphicHeight}
\newcommand{\mainfigurepage}[4]{%
  \clearpage
  \newgeometry{left=10mm,right=10mm,top=10mm,bottom=17mm,footskip=9mm}%
  \thispagestyle{plain}%
  \begin{lrbox}{\MainCaptionBox}%
    \begin{minipage}{\linewidth}%
      \captionsetup{type=figure,font={small,stretch=1.05},labelfont=bf,labelsep=bar,name=Fig.}%
      \caption{\textbf{#3.} #4}\label{fig:main#1}%
    \end{minipage}%
  \end{lrbox}%
  \setlength{\MainGraphicHeight}{\dimexpr\textheight-\ht\MainCaptionBox-\dp\MainCaptionBox-24pt\relax}%
  \noindent\makebox[\linewidth][c]{\includegraphics[width=\linewidth,height=\MainGraphicHeight,keepaspectratio]{#2}}\par
  \vspace{2mm}%
  \noindent\usebox{\MainCaptionBox}\par
  \clearpage
  \restoregeometry
}

\newcommand{\ManuscriptTitle}{Evidence-grounded large language models guide perovskite solar cell design and optimization}
\newcommand{\AuthorLine}{Chen YanKai\textsuperscript{1,\ensuremath{\dagger}}, Wan Zhi\textsuperscript{2,\ensuremath{\dagger},*}, Li ZiHao\textsuperscript{3,\ensuremath{\dagger}} and Jing Tao\textsuperscript{1,*}}
\newcommand{\AffiliationBlock}{%
\textsuperscript{1}School of Materials Science and Engineering, Tsinghua University, Beijing~100084, P.~R.~China.\\
\textsuperscript{2}College of Bioresources Chemical and Materials Engineering, Shaanxi University of Science \& Technology, Xi'an~710021, P.~R.~China.\\
\textsuperscript{3}Department of Chemical Engineering, Tsinghua University, Beijing~100084, P.~R.~China.}
\newcommand{\AuthorNotes}{%
\textsuperscript{\ensuremath{\dagger}}These authors contributed equally.\\
\textsuperscript{*}Corresponding authors: Wan Zhi and Jing Tao.\\
Contact e-mails: \href{mailto:wanzhi@sust.edu.cn}{\texttt{wanzhi@sust.edu.cn}} (Wan Zhi); \href{mailto:jingtao@mail.tsinghua.edu.cn}{\texttt{jingtao@mail.tsinghua.edu.cn}} (Jing Tao).}
\newcommand{\PDFAuthorNames}{Chen YanKai; Wan Zhi; Li ZiHao; Jing Tao}

\hypersetup{hidelinks,pdftitle={\ManuscriptTitle},pdfauthor={\PDFAuthorNames}}
\begin{document}
\setstretch{1.15}
\renewcommand{\bibsection}{\section*{References}}
\begin{center}
  {\LARGE\bfseries \ManuscriptTitle\par}
  \vspace{0.9em}
  {\large \AuthorLine\par}
  \vspace{0.65em}
  {\small\setstretch{1.1}\AffiliationBlock\par}
  \vspace{0.6em}
  {\footnotesize\setstretch{1.1}\AuthorNotes\par}
\end{center}
\begin{abstract}
Using large language models to interpret experimental evidence, design discriminating experiments and formulate well-supported mechanistic explanations remains a central challenge in materials research. Here we establish PSM-Bench, an 800-question benchmark developed by perovskite researchers to evaluate scientific reasoning in perovskite photovoltaics. We also introduce OpenPerov, an expert-informed language-model framework that learns from expert-calibrated evidence records reconstructed from 4,035 training articles; these records link experimental strategies, controls and observations to mechanistic interpretations. Structured-evidence adaptation outperformed conventional question--answer training, and OpenPerov Flash ranked first among 13 closed-book systems. OpenPerov Pro raised Scientific Mastery further, from 87.6 to 89.1, by revising selected passages with retrieved evidence. On questions derived from subsequent studies, both models also outscored expert-authored answers. In an experimental test of self-assembled monolayer design, OpenPerov selected a K analogue that outperformed the Na analogue preferred by experts. Devices incorporating the selected contact reached a power conversion efficiency of 27.05\% and retained 96.74\% of their initial efficiency after 1,000 h of dark thermal ageing at 85 $^\circ$C. These results show that reconstructed scientific reasoning provides an effective basis for building domain models that interpret experiments and inform materials design.

\end{abstract}
\section*{Introduction}

Large language models (LLMs) increasingly assist scientific research, bringing domain knowledge to bear on hypothesis generation and experimental design \citep{Boiko2023Coscientist,Marwitz2026Concept}. In materials science, literature mining and domain-specific LLMs now support candidate screening, synthesizability assessment and precursor selection \citep{Tshitoyan2019Latent,Dagdelen2024Structured,Li2026Reconstruction,Ahlawat2026LLaMat,Tang2026MatterChat,Song2025CSLLM}. In perovskite photovoltaics, their applications range from performance prediction and additive selection to experimental optimization \citep{Xie2024StructuredPV,Liu2025PerovskiteLLM,Wang2026PerovskiteR1,Sun2026PVKLLM}. Building on these advances, AI-guided design of molecular contacts and dopants, combined with automated screening and fabrication, applies computational recommendations to photovoltaic device development and optimization \citep{Guo2026ScienceAI,Wang2026HTLDopant,Gao2026AutonomousPV}. This growing role in research practice demands stronger scientific reasoning from domain models: linking observations to mechanisms and designing discriminating experiments under coupled material and operating constraints.

Here we establish PSM-Bench to evaluate scientific reasoning in perovskite photovoltaics and introduce OpenPerov to develop this capability. PSM-Bench comprises 800 expert-developed questions drawn from 446 studies. Its criteria link mechanistic explanations to observable consequences and experimental controls, and they accommodate alternative answers that account for the evidence. OpenPerov Flash learns from expert-calibrated evidence records reconstructed from 4,035 training articles within a curated corpus of 4,532 articles. These records link research questions and experimental strategies to observations, mechanistic interpretations and literature lineage. Organized in this way, the records preserve the conditions and comparisons through which evidence supports an explanation. OpenPerov Pro retrieves relevant evidence from articles and their citation context, and uses it to refine selected parts of an answer \citep{Lewis2020RAG,Gao2023RARR}.

Flash ranked first among 13 closed-book systems on PSM-Bench, a lead corroborated by double-blind expert assessment. Compared with conventional question--answer adaptation on the same articles and backbone, the structured-evidence learning route performed better with fewer training tokens and kept its advantage after identical subsequent training. Targeted evidence revision delivered larger gains than regenerating complete answers from the same retrieved material. To test the framework on a complex interfacial system, we applied it to the design of self-assembled monolayers (SAMs), in which molecular packing, anchoring and binding geometry jointly determine performance \citep{AlAshouri2019Conformal,Zhao2024Peri,Jiang2025Toughened,Xiong2026Electrodeposition,Yang2026Stereoelectronic}. OpenPerov proposed a molecular scaffold that combines oxide anchoring with perovskite-facing carboxylate groups, and prioritized BCPCz--TMPS--K as its preferred candidate. Under matched fabrication conditions, this candidate outperformed the Na analogue favoured by human experts.

\section*{Results}

We first introduce PSM-Bench, a comprehensive benchmark for evaluating large language models in perovskite photovoltaics and related research. It assesses the reasoning that links scientific observations to experimental decisions, research designs and the interpretation of findings. Perovskite experts and their research teams developed its 800 open-ended questions from 446 studies, covering mechanism diagnosis, stability-constrained design, stability-failure analysis and cross-constraint synthesis (Fig.~\ref{fig:main2}a and Supplementary Fig. S6). Together, these tasks capture the central demands of perovskite research: explaining the physicochemical origins of experimental observations, distinguishing competing mechanisms, and selecting interventions for multidisciplinary problems under coupled material, device and operating constraints. Most questions call for several kinds of reasoning at once; 78.5\% combine at least three requirements spanning experimental discrimination, mechanistic explanation, decision logic, and alternatives or applicability conditions.

The interface-design question in Fig.~\ref{fig:main2}a illustrates the coupling between chemical passivation and charge transport. Increasing the modifier dose strengthens XPS signatures associated with passivation, yet lowers the fill factor and raises contact resistance; the chemical improvement thus comes with an electrical penalty. Greater coverage may passivate more sites, but it may also alter the thickness, connectivity or energetics of the contact, and each possibility calls for a different corrective action. Linking these alternatives to the outcomes they predict in discriminating measurements connects mechanistic theory with experimental evidence. PSM-Bench evaluates whether a model can carry out this kind of expert reasoning, from interpreting observations to choosing the next experiment and design intervention.

Each PSM-Bench question carries four scientific criteria, acceptable alternative answers and specified reasoning errors. Its central metric, Scientific Mastery, combines semantic adequacy, rubric coverage, supported-claim precision and decision or control quality on a 0--100 scale (Methods; Supplementary Figs. S7--S9). The criteria follow the structure of the scientific argument, linking a model's proposed explanation to its observable consequences. The scoring therefore accommodates alternative mechanisms while distinguishing how fully and precisely each accounts for the evidence and motivates experimental choices.

OpenPerov Flash ranked first among the 13 open- and closed-source systems evaluated on PSM-Bench under closed-book conditions. It reached a Scientific Mastery score of 87.6, compared with 83.3 for the strongest comparator, GLM-5.2 (Fig.~\ref{fig:main2}c). Flash led all four task families and outscored every comparator on a majority of questions, including 63\% of questions against GLM-5.2 (Fig.~\ref{fig:main2}d and Supplementary Fig. S16). Its advantage thus held both across different forms of scientific reasoning and across individual problems. We further developed OpenPerov Pro, which uses the same answer-model checkpoint to revise selected passages with retrieved evidence and raises Scientific Mastery to 89.1.

The largest margins over GLM-5.2 arose in stability-constrained design and cross-constraint synthesis, the more difficult task families (Supplementary Figs. S7 and S15). These tasks require a model to choose among interventions whose benefits depend on several coupled conditions. A treatment that improves initial efficiency, for example, must also be evaluated in relation to the processes that govern performance under stress. Among the scoring components, Flash's clearest advantage lay in supported-claim precision. These results show that Flash's strength lies in making scientifically grounded choices under interacting constraints and translating them into well-supported, feasible experimental strategies.

Flash's lead held under expert reassessment and with alternative prompts (Supplementary Figs. S10 and S11). In a double-blind assessment, volunteer perovskite experts evaluated 3,200 anonymized answers from four systems, producing 12,800 criterion-level ratings. Flash ranked first in 99.84\% of human-score bootstrap resamples. This agreement between expert ratings and the benchmark ranking independently validates Scientific Mastery as a measure of scientific answer quality.

To develop this capability, we organized the literature around the arguments through which scientific findings are established. Domain experts reconstructed and calibrated evidence from 4,532 articles, integrating the main text, methods, supporting information, figures and tables (Fig.~\ref{fig:main1}a,b). Each Integrated Article Evidence Record connects a research question to the experimental strategy, controls, observations and mechanistic interpretation under the relevant conditions. A measurement is thus represented together with the explanation it supports and the alternatives that the experiment addresses. This organization preserves the relationships needed to move from an observation to further characterization, mechanistic analysis and experimental decisions.

A record of light-induced iodine redistribution illustrates the scientific content that this reconstruction recovers (Supplementary Fig. S5). Photoluminescence brightening coincides with an approximately tenfold reduction in fitted trap density. The brightening persists in nitrogen and in vacuum, establishing an oxygen-independent response, while composition maps reveal lateral iodine depletion and enrichment as well as redistribution through the film thickness. These observations motivate a trap--ion coupling hypothesis: trap filling generates local fields that promote iodide redistribution and alter the population of electronically active defects. The atmospheric controls constrain the chemical explanation, and the spatial maps tie the optical change to ionic motion. By retaining these relationships, the record presents the proposed mechanism together with the observations and controls from which it arose.

We constructed a systematic literature lineage to connect these arguments across perovskite studies (Supplementary Figs. S1--S4). Earlier findings supply the hypotheses and methods that motivate an experiment; later studies develop, qualify or revise its interpretation. A single article can therefore contribute a successful design, a discriminating control or an unresolved mechanistic question. Observations from figures and tables enter these records as source-linked scientific text that retains their experimental role. The resulting corpus captures not only the findings but also the research choices that produced them. Flash learns these relationships during adaptation, whereas Pro uses the corresponding evidence records to revisit specific parts of an answer when additional information becomes relevant.

Flash was adapted from Qwen3.6-27B \citep{Qwen2026Backbone} with LoRA through domain-adaptive pretraining (DAPT), targeted scientific instruction training and answer-style tuning \citep{Hu2022LoRA,Gururangan2020DAPT,Wei2022FLAN} (Fig.~\ref{fig:main1}c). The adaptation set comprised 4,035 articles. The remaining 497 were held out, and content overlapping with their key scientific conclusions was excluded from project-generated training data. To test the learning route, we compared structured-evidence DAPT with conventional supervised learning from PDF-derived question--answer records, using the same source articles and backbone. The structured corpus contained 78,299 records and 19.33 million tokens, whereas the conventional corpus contained 211,518 records and 63.39 million tokens. The structured learning route yielded stronger scientific reasoning than conventional instruction--answer adaptation despite using fewer training tokens.

After initial adaptation, the structured route reached a Scientific Mastery score of 82.3, compared with 71.5 for the conventional route and 75.8 for the unadapted backbone (Fig.~\ref{fig:main2}b). This corresponded to a 10.77-point advantage over conventional adaptation. Both checkpoints then received identical scientific instruction and answer-style training, so that the two initial learning routes could be compared after the same subsequent adaptation. The structured route retained a 3.50-point advantage, reaching 87.57 versus 84.07, with gains in all four task families (95\% confidence interval for the paired difference, 2.60--4.34; Supplementary Fig. S12). The initial learning route thus continued to shape the model's final capability even after shared targeted training.

Ablations resolve the complementary contributions of the training stages (Supplementary Fig. S13). Removing DAPT reduced Scientific Mastery by 1.59 points, whereas omitting both the subsequent scientific instruction and answer-style training stages reduced it by 5.26 points. DAPT benefited every task family. The later stages had their largest effects on stability-constrained design and cross-constraint synthesis, where an answer must turn several scientific relationships into a coherent choice. This pattern connects learning the structure of research evidence with learning to apply that structure to an experimental problem. Adaptation also improved every model scale tested, from 8B to 32B; the 27B and 32B models showed similar gains, and Qwen3.6-27B gave the strongest final performance (Supplementary Fig. S14).

We next examined how additional evidence could improve reasoning once an initial answer had been formed. Pro retrieves from 173,024 article-centred documents spanning the full collection (Fig.~\ref{fig:main3}a). Knowledge extraction retains each observation together with its source, experimental conditions and scientific role; Ranking Compression prioritizes relevant articles and assembles the evidence that the question requires. Source selection raised retention of the question's originating article from 82.75\% to 91.00\%, and scientific-relevance reranking improved nDCG@10 from 0.8107 under fixed fusion to 0.8739 (Supplementary Fig. S18) \citep{Zhang2025Qwen3Embedding}. The two operations meet distinct needs: recovering the relevant study, and bringing the most applicable evidence to the front. Because compression preserves conditions and complete scientific statements, the answer model has access to the relationships on which a revision depends.

Pro uses this evidence to identify an unresolved scientific requirement in the Flash answer and to address it through a local addition or replacement. The unit of revision is the part of the argument that new evidence can improve: a missing causal link, an insufficiently discriminating control, or a prediction that needs a relevant experimental condition. Each change is linked to supporting evidence, and the surrounding reasoning is retained. This distinction matters when an answer already connects several observations correctly, because a useful new finding may resolve only one remaining ambiguity in the argument.

A matched comparison with full-answer retrieval-augmented generation (RAG) revealed this effect in the distribution of answer changes \citep{Lewis2020RAG} (Fig.~\ref{fig:main3}b). With the same 800 questions, answer-model checkpoint and evidence packets, full-answer RAG produced score gains of at least five points on 25.75\% of questions but losses of the same magnitude on 24.38\%. The overall Scientific Mastery gain was 0.22 points. Pro produced substantial gains on 15.75\% of questions and substantial losses on 0.50\%, for an overall gain of 1.49 points (Supplementary Fig. S19). Full-answer regeneration generated more large improvements alongside frequent large declines. Local revision shifted the distribution towards preserving the quality of the initial answer while incorporating useful evidence, which accounts for its larger aggregate improvement.

We compared Pro's acceptance of all structurally valid revisions with a critic-selected alternative. Critic-selected revision yielded a 1.23-point gain with the same 0.50\% substantial-loss rate, but reduced the frequency of substantial gains to 12.75\% (Fig.~\ref{fig:main3}b). Both strategies maintained a low rate of large declines, while Pro captured more improvements.

A mixed-halide example shows what such an improvement means scientifically. The question combined a weaker photoluminescence redshift with continued halide migration, AgI formation at the metal contact and hysteresis. Pro added a causal connection through carrier extraction, which can weaken the optical redshift while ionic redistribution and contact reactions continue (Supplementary Methods SM6.2). The revised explanation assigns different sensitivities to the observables: luminescence reflects the distribution and recombination of photocarriers, whereas contact chemistry and hysteresis can reveal ongoing ionic processes. This connection reconciles apparently conflicting measurements and changes how the reduced redshift is interpreted. The optical improvement can coexist with degradation elsewhere in the device, which directs attention to measurements that resolve both processes.

We then tested scientific reasoning on ten studies published online after the literature-collection cutoff of 14 April 2026 (Fig.~\ref{fig:main3}c and Supplementary Table 1). Each study supplied four tasks in each of two information settings. The experimental-context setting provided observations for mechanism diagnosis, decisive-experiment selection, evidence updating and scientific decisions. The limited-context setting required competing explanations and falsifiable predictions from sparse observations. Together, the two settings examine how a model uses the measurements at hand and how it plans to obtain missing information. Flash and Pro answered 40 questions in each setting without access to the target articles or their supporting information. Experts answered the same questions with access to online resources available up to the same literature cutoff, reflecting routine research practice. Independent volunteer perovskite experts evaluated the anonymized responses using setting-specific rubrics (Methods).

Under the experimental-context rubric, Pro scored 88.2, Flash 87.2 and the expert-authored answers 85.5. In the limited-context setting, designed to reflect research decisions made from sparse observations, the separate rubric yielded scores of 96.8, 95.8 and 92.9, respectively (Fig.~\ref{fig:main3}c and Supplementary Fig. S20). The largest improvements concerned decisive experiments and scientific decisions when observations were available, and discrimination between competing mechanisms when information was sparse. These differences reveal where evidence revision contributes most directly to research: selecting the measurement that changes the next decision, and specifying the observation that separates plausible explanations. A complementary external assessment on the independently released Perovskite-R1 benchmark gave 46 correct answers out of 49 for Flash and 48 for Pro, compared with 45 in the authors' released predictions \citep{Wang2026PerovskiteR1} (Supplementary Fig. S23). OpenPerov's advantage thus extends to a focused external test of perovskite knowledge.

In the substrate-side deep-UV task (Fig.~\ref{fig:main3}d), Flash proposed a dose series under the half-dose constraint. Pro incorporated substrate transmission and wavelength-dependent absorption to focus on the photon fluence reaching the modifier layer, and to identify activation thresholds, saturation and damage. Under limited context, Pro paired a terminal-alkene-free control with depth-resolved or optically calibrated reaction profiles to distinguish alkene-dependent photochemistry from general passivation and nonuniform activation, linking dose optimization to mechanistic discrimination.

We applied this reasoning to the design of a solution-processable, hole-selective SAM at the hydroxylated indium tin oxide (ITO)/perovskite interface (Fig.~\ref{fig:main4}a,b). The design brief combined oxide anchoring, monolayer integrity, Pb passivation and control of interfacial ions, while seeking an alternative to phosphonic- or phosphoric-acid anchoring \citep{Fei2026AcidReactivity}. Introducing evidence for trialkoxysilane attachment to hydroxylated oxides and for carboxylate interactions with undercoordinated Pb guided the design towards separate oxide-facing and perovskite-facing functions. Flash selected a trimethoxysilyl branch at the carbazole N position and two carboxylate branches at positions 3 and 6, and Pro refined the proposal through evidence-conditioned revision. The design thus evolved from an initial chemical concept into a molecule whose substituents address specific requirements of the buried contact.

In the resulting BCPCz--TMPS--M architecture (M = Na or K), a para-phenylene spacer links carbazole N9 to a trimethoxysilyl group, and further para-phenylene spacers link C3 and C6 to two carboxylate groups. The proposed interfacial mechanism couples oxide attachment to coordination at the perovskite surface. Hydrolysis of the trimethoxysilyl group supplies silanols for condensation with the hydroxylated oxide, following the established anchoring chemistry of silane-based hole-transport SAMs \citep{Tang2024Reinforcing}. Oxide attachment and limited lateral condensation stabilize coverage, while the two carboxylate branches provide multiple sites for interaction with undercoordinated Pb. This division of chemical functions integrates substrate anchoring and perovskite passivation into the hole-selective carbazole framework, with the aim of preserving contact integrity and regulating the interfacial electronic and ionic environment.

Counterion selection extends this mechanism to the balance between coordination and ionic association. Flash chose K after comparing Li, Na and K, whereas the experts favoured Na. With its larger ionic radius and lower charge density, K shows distinct electrostatic and solvation behaviour around the carboxylate groups. Counterion-dependent ion pairing tunes the competition between alkali-metal association and Pb coordination at the carboxylate oxygen: the more diffuse electrostatic interaction around K favours access of undercoordinated Pb to the carboxylate groups, while the counterion modifies charge compensation and halide distribution near the contact. Potassium passivation has previously reduced non-radiative losses and photoinduced ion migration in perovskite films and interfaces \citep{AbdiJalebi2018Potassium}, providing a chemical rationale for examining K at the buried contact. Introducing the counterion through the same molecular scaffold allows its contribution to be tested by changing the alkali metal while retaining the anchoring and coordinating architecture.

Pro retained K and translated the proposed functions into comparisons capable of distinguishing their contributions (Supplementary Methods SM9). It specified a carboxylic-acid or no-salt control, matched Li/Na/K analogues, and measurements separating attachment to ITO from lateral siloxane condensation. It also proposed tests of carboxylate--Pb coordination, K redistribution, interface energetics and charge extraction. These choices address the balance inherent in the molecule: anchoring must preserve coverage, while condensation and ionic association must still permit an effective electronic contact. The experimentally decisive comparison was between the expert-preferred Na analogue and the model-selected K analogue. We synthesized both and used identical SAM preparation, deposition and subsequent device-fabrication procedures.

The K analogue increased the mean power conversion efficiency (PCE) by 1.45 percentage points. Across independent groups of 16 devices each, BCPCz--TMPS--K (SAM-K) reached 26.69 $\pm$ 0.27\%, compared with 25.24 $\pm$ 0.30\% for BCPCz--TMPS--Na (SAM-Na; mean $\pm$ s.d.; Supplementary Fig. S24 and Supplementary Table 2). The improvement was most pronounced in the fill factor, which rose from 83.28\% to 85.71\%, accompanied by an increase in open-circuit voltage from 1.174 to 1.191 V. The short-circuit current density increased from 25.82 to 26.15 mA cm$^{-2}$. The concurrent voltage and fill-factor gains are consistent with reduced recombination losses and improved charge extraction at the modified contact. The measured gains establish counterion selection as an experimentally consequential element of molecular design.

SAM-K also outperformed the conventional Me-4PACz contact. In small-area Ag-electrode cells measured through a 0.09 cm$^2$ aperture, SAM-K reached a reverse-scan PCE of 27.05\%, compared with 25.53\% for Me-4PACz (Fig.~\ref{fig:main4}c). Fixed-bias measurements over approximately 600 s gave stabilized power-output efficiencies of 26.8\% and 24.68\%, respectively (Fig.~\ref{fig:main4}e), and the EQE-integrated short-circuit current densities were 25.60 and 25.26 mA cm$^{-2}$ (Fig.~\ref{fig:main4}d). The modest difference in photocurrent, set against the efficiency gain under both scanning and fixed-bias measurement, is consistent with the contact-quality interpretation of the Na/K comparison. The designed interface supports efficient power conversion while preserving carrier collection.

We assessed whether this performance persisted under three ISOS-based ageing protocols, using encapsulated devices in ambient air \citep{Khenkin2020ISOS}. Modified ISOS-LC-2 cycling alternated 12 h of illumination at 85 $^\circ$C under maximum-power-point (MPP) tracking with 12 h in the dark at approximately 20 $^\circ$C, repeatedly changing temperature, photocarrier generation and electrical operation. SAM-K retained 89.8\% of its first complete illuminated-cycle efficiency after 1,037.5 h. At the common comparison time of approximately 534 h, its normalized efficiency exceeded that of Me-4PACz by approximately 10.2 percentage points (Fig.~\ref{fig:main4}f). The advantage thus persisted through repeated transitions between illuminated operation and dark cooling.

Separate thermal and operational tests probed performance under sustained stress. During ISOS-D-2 dark ageing at 85 $^\circ$C, SAM-K retained 96.74\% of its initial efficiency after 1,000 h; Me-4PACz retained 70.6\% after 730 h (Fig.~\ref{fig:main4}g). Under ISOS-L-1 continuous one-sun-equivalent illumination with MPP tracking, SAM-K and Me-4PACz retained 94.84\% and 81.58\%, respectively, of their initial efficiencies after 994 h (Fig.~\ref{fig:main4}h). Across thermal, continuous-operation and cycling stresses, these results support the design objective of sustaining an effective buried contact. Together with the voltage and fill-factor gains, these results are consistent with the proposed coupling of contact integrity and interfacial electronic and ionic regulation.

\section*{Conclusion}

OpenPerov leads PSM-Bench across all of its task families, a result corroborated by double-blind expert assessment. Its performance on questions from subsequent studies extends this capability to both experimental-context and limited-context settings. The framework treats scientific reasoning as both the substance of domain adaptation and the standard of evaluation, linking expert-calibrated learning with targeted evidence revision to formulate decisive experiments under complex constraints.

In SAM molecular design, OpenPerov proposed BCPCz--TMPS--K, which assigns substrate anchoring and perovskite passivation to distinct chemical units. Its design rationale couples coordination and ion pairing at the perovskite surface to regulation of the electronic and ionic environment, while anchoring maintains contact integrity. Experimental validation establishes this architecture as a practical strategy for SAM interface engineering. Across the evaluated design tasks, OpenPerov demonstrated stronger experimental-design reasoning than the participating human experts, showing its potential to advance materials discovery through the coordinated design of molecules, controls and measurements.

\mainfigurepage{1}{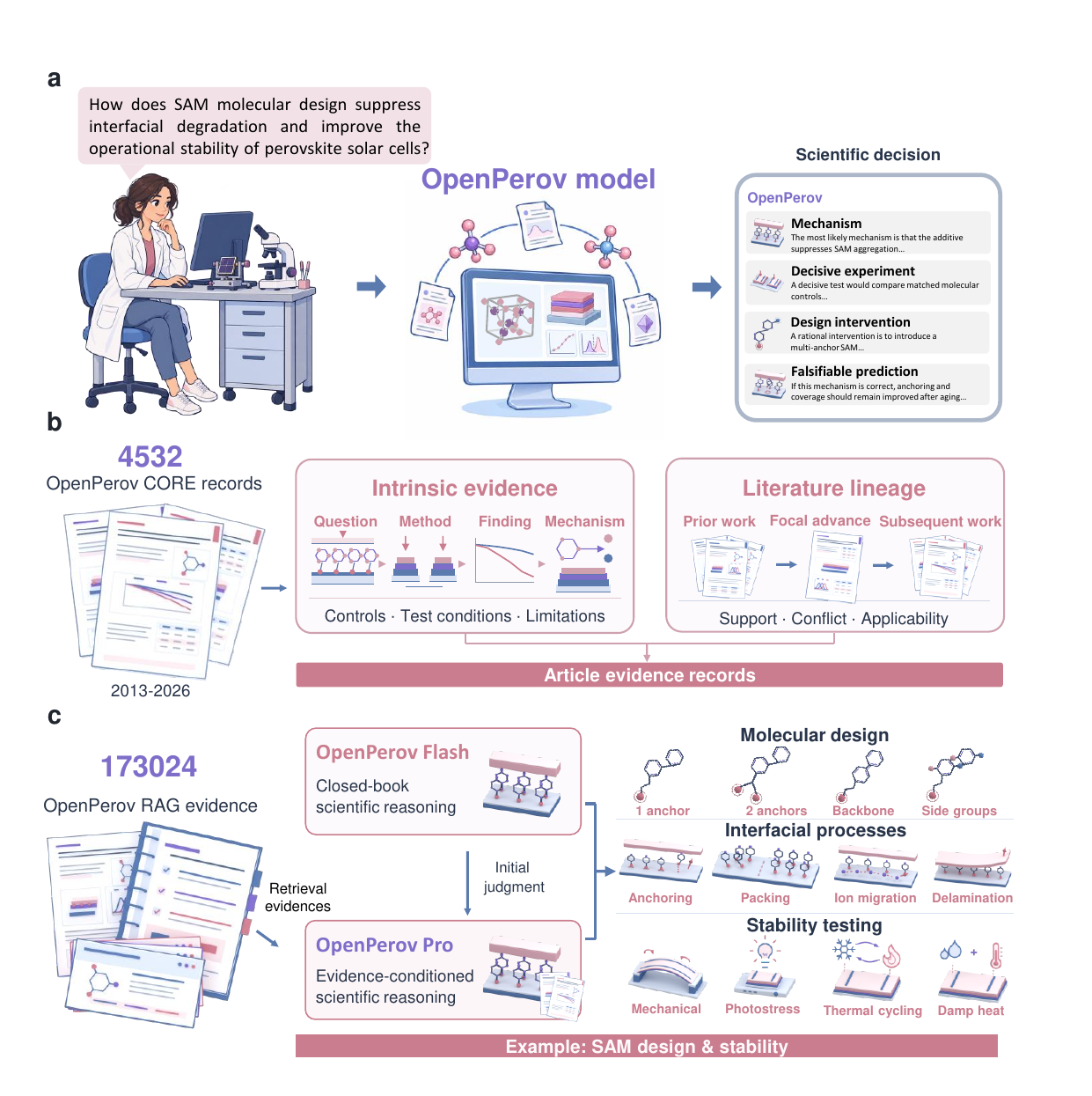}{Article-centred evidence connects literature knowledge to scientific reasoning in OpenPerov}{\textbf{a,} Overview of the framework, illustrated with a self-assembled monolayer (SAM) stability question that links mechanistic interpretation to a discriminating experiment, a design intervention and a prediction. \textbf{b,} Evidence from 4,532 curated articles is organized into records that connect research questions, controls, observations and interpretations with experimental conditions and selected literature lineage. \textbf{c,} After domain adaptation, OpenPerov Flash answers without retrieval, whereas OpenPerov Pro retrieves relevant evidence to revise an initial Flash answer. Figure and table evidence is represented as article-linked scientific text (Supplementary Methods SM1--SM3 and SM6).}

\mainfigurepage{2}{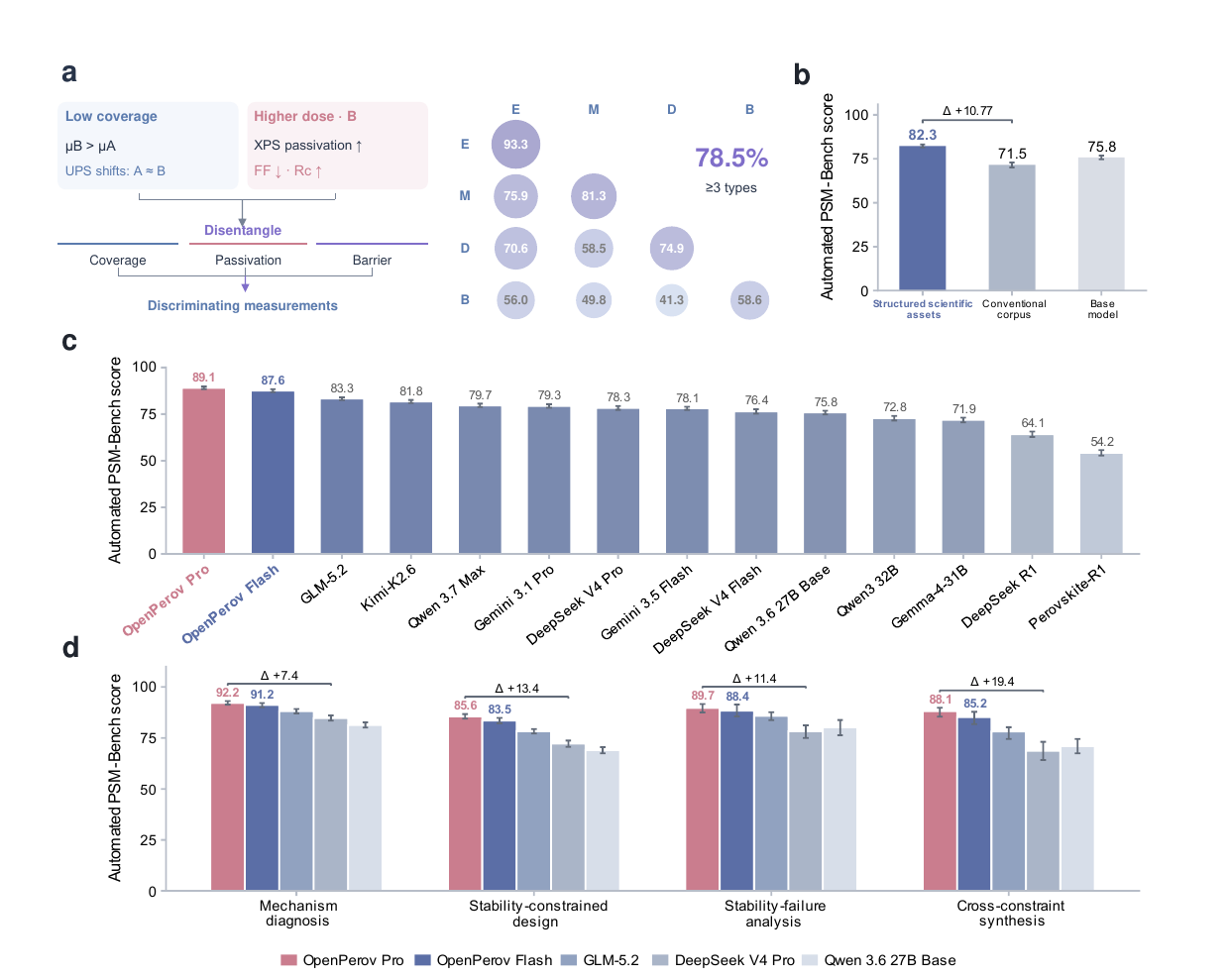}{Scientific reasoning evaluated with PSM-Bench}{\textbf{a,} An illustrative question and the reasoning requirements across the 800 expert-developed items. The example separates passivation-related from transport-related explanations through discriminating measurements. Matrix diagonals show the prevalence of experimental discrimination (E), mechanistic explanation (M), design or decision logic (D), and alternatives or applicability boundaries (B); off-diagonal entries show pairwise co-occurrence. At least three requirements co-occur in 78.5\% of items. \textbf{b,} Two initial learning routes using the same backbone were compared: structured scientific evidence and conventional PDF-derived instruction--answer text, both drawn from the same 4,035 source articles. After initial adaptation, the routes scored 82.3 and 71.5, respectively, against 75.8 for the unadapted backbone. The 10.77-point difference is calculated from unrounded scores. Both checkpoints then received the same scientific instruction and answer-style training (Supplementary Fig. S12). \textbf{c,} Overall Scientific Mastery scores for 13 closed-book systems and for evidence-conditioned OpenPerov Pro. \textbf{d,} Scores across the four task families for representative systems. Brackets compare Pro with DeepSeek V4 Pro. Scores use Huber aggregation on a 0--100 scale; error bars indicate 95\% item-bootstrap confidence intervals (SM3--SM5).}

\mainfigurepage{3}{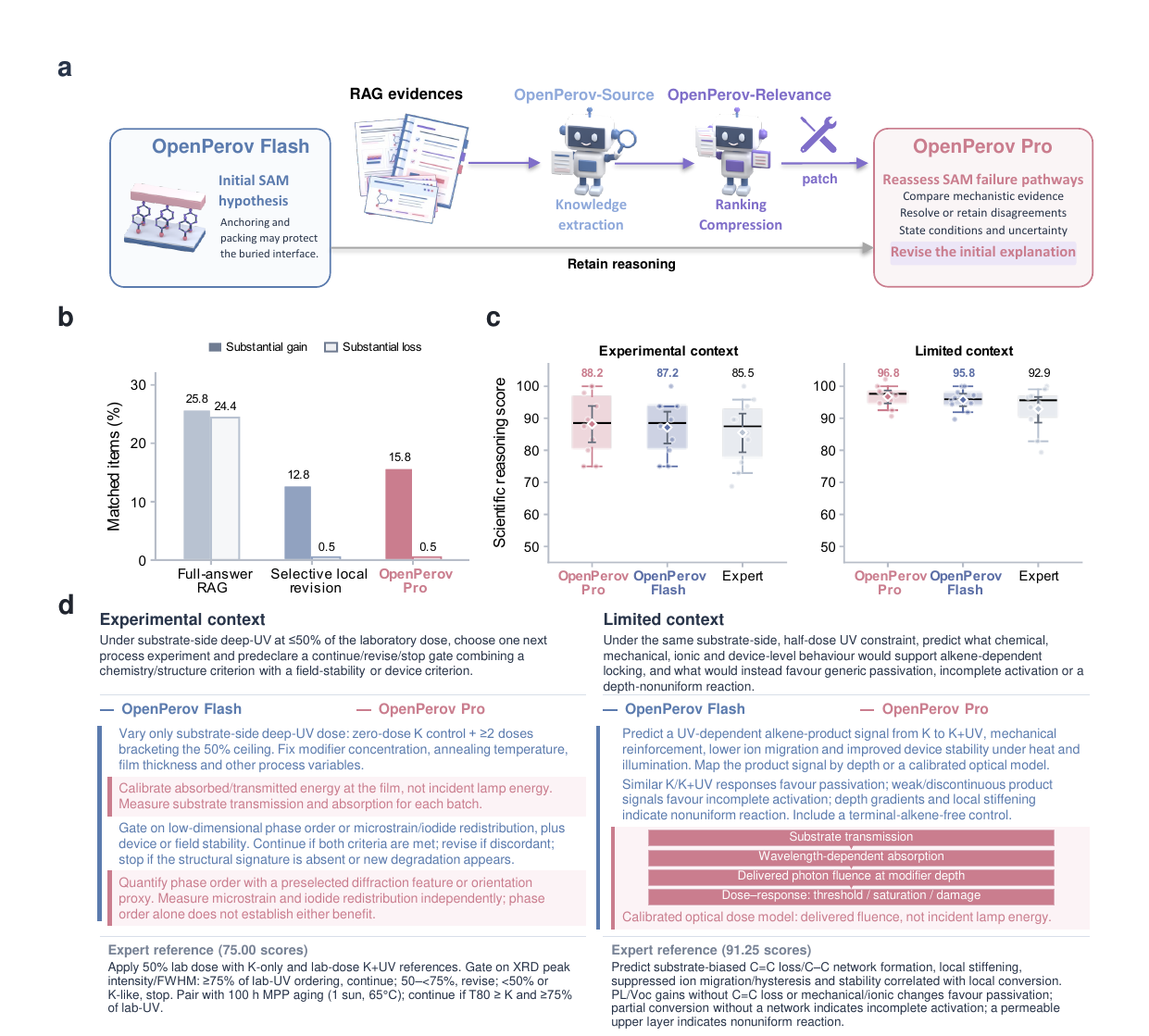}{Evidence-linked revision supports scientific reasoning}{\textbf{a,} OpenPerov Pro uses an initial Flash answer and retrieved evidence to make local additions or replacements. Knowledge extraction retains source-linked evidence; Ranking Compression orders relevant articles and assembles their content within the evidence budget (SM6--SM7). \textbf{b,} Effects on 800 matched PSM-Bench questions relative to Flash. Substantial gains and losses are score increases and decreases of at least 5 points, respectively. Gain/loss frequencies are 25.75\%/24.38\% for full-answer RAG, 12.75\%/0.50\% for critic-selected revision and 15.75\%/0.50\% for the Pro setting that applies all structurally valid revisions. \textbf{c,} Ten post-cutoff studies supplied 40 questions per information setting. Pro/Flash/Expert means are 88.2/87.2/85.5 with experimental context and 96.8/95.8/92.9 with limited context. Each setting has its own tasks and rubric. When preparing their answers, experts could consult online resources available up to OpenPerov's literature cutoff, reflecting routine research practice (SM7). Points show study means across four questions. Boxes show interquartile ranges, centre lines show medians, and whiskers span the 5th--95th percentiles. Diamonds show means with 95\% study-cluster bootstrap intervals. \textbf{d,} Example experimental decisions and predictions under a substrate-side deep-UV constraint. Pro emphasizes the photon exposure delivered at the depth of the modifier. This evaluation used task-specific acceptance criteria and at most two edits per answer (SM7).}

\mainfigurepage{4}{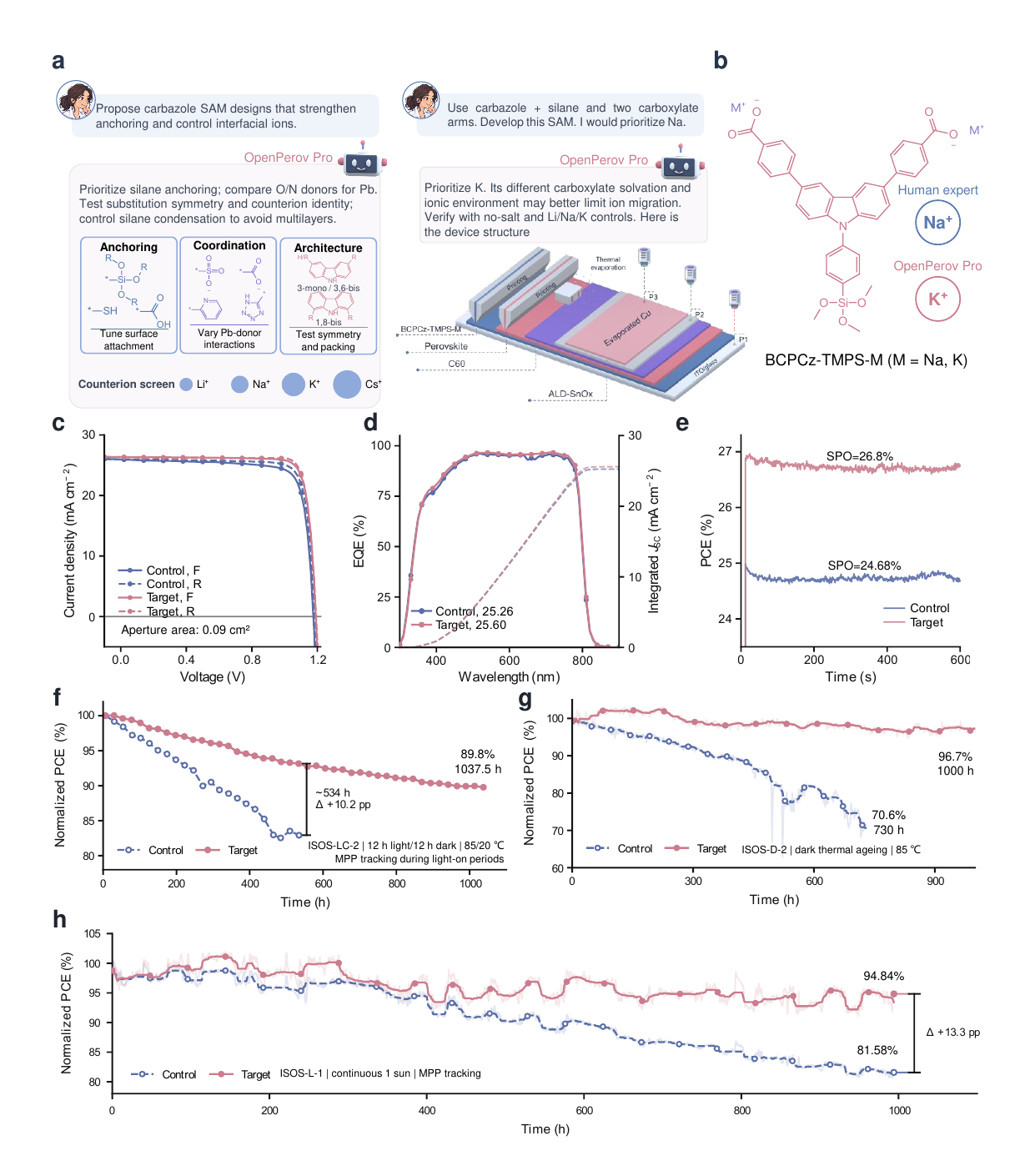}{Experimental comparison of expert and model choices for SAM design}{\textbf{a,b,} Schematic of the design rationale for BCPCz--TMPS--M (M = Na or K), contrasting the human experts' preference for Na with OpenPerov's choice of K. Devices incorporating the model-selected K analogue achieved a higher mean efficiency than those incorporating Na (Supplementary Fig. S24 and Supplementary Table 2). SM9 describes the sequence of Flash proposals and Pro revisions. The illustrated module uses a Cu electrode and P1/P2/P3 laser patterning. \textbf{c--h,} Measurements on small-area Ag-electrode cells comparing the conventional Me-4PACz SAM (Control) with BCPCz--TMPS--K (Target). \textbf{c--e,} Current-density--voltage curves, external quantum efficiency spectra and fixed-voltage power output recorded for approximately 600 s, respectively. Reverse-scan efficiencies are 25.53\% for Control and 27.05\% for Target. \textbf{f,} Light--dark cycling under a modified ISOS-LC-2 protocol, with repeated 12 h illuminated periods at 85 $^\circ$C under MPP tracking, each followed by a 12 h dark period at approximately 20 $^\circ$C with tracking stopped. Markers show mean efficiencies over complete illuminated periods. \textbf{g,} Dark thermal ageing at 85 $^\circ$C following ISOS-D-2. Target retained 96.74\% after 1,000 h, and Control retained 70.6\% after 730 h. \textbf{h,} Continuous one-sun-equivalent illumination with MPP tracking under ambient-temperature ISOS-L-1 conditions. After 994 h, Target retained 94.84\% and Control 81.58\%, a difference of 13.26 percentage points. Cells in f--h were encapsulated and tested in ambient air. In g and h, efficiencies are normalized to each group's initial value (SM13). Pale traces show measured values; rolling medians guide the eye (SM10--SM14).}

\section*{Methods}

\subsection*{Scientific corpus and model adaptation}

The corpus contains 4,532 articles on perovskite photovoltaics published in 2013--2026. Perovskite-domain experts reconstructed each study's research question, experimental approach, controls, observations and interpretation from its main text, supporting information, figures and tables, calibrating the scientific content article by article. GPT-5.5 and GPT-5.6 Sol assisted with text processing of these expert-constructed records. Selected earlier and later studies provided context, and figures and tables were represented as source-linked scientific text. We assigned 4,035 articles to model adaptation and excluded the remaining 497 from project-generated training data. We also removed training content that overlapped with the key scientific conclusions of the held-out studies, including overlap conveyed through references or citing papers. The retrieval collection covered all 4,532 articles (Supplementary Methods SM1--SM2).

OpenPerov Flash was adapted from Qwen3.6-27B \citep{Qwen2026Backbone} through DAPT, targeted scientific instruction training and answer-style tuning \citep{Gururangan2020DAPT,Wei2022FLAN}. At each stage, we fitted and merged a LoRA adapter with rank $r=16$, scaling parameter $\alpha=32$ and dropout 0.05 \citep{Hu2022LoRA}. DAPT used 78,299 structured records, 1,024-token blocks and a learning rate of 5 $\times$ 10$^{-6}$. The two subsequent stages used 43,840 scientific instruction examples and 7,490 answer-style examples, respectively, with supervision applied to assistant-answer tokens. The conventional-learning route used assistant-only causal-language-model training on 211,518 PDF-derived instruction--answer records from the same 4,035 source articles. The two routes shared the backbone, context length, LoRA configuration and subsequent training stages, but retained their respective data representations and initial objectives. SM3 describes the training stages, learning-route comparison and component ablations.

\subsection*{Scientific tasks and expert assessment}

PSM-Bench comprises 800 open-ended questions grounded in 446 studies, covering mechanism diagnosis, stability-constrained design, stability-failure analysis and cross-constraint synthesis. Experts developed and finalized the scientific content and scoring criteria. All closed-book systems received the frozen questions without source passages or reference answers, and only their final answers were scored. Scientific Mastery combines semantic adequacy, rubric coverage, supported-claim precision and decision or control quality through a deterministic item-score calculation followed by Huber aggregation \citep{Huber1964Robust}. SM4--SM5 define the calculation and inference settings. In a double-blind assessment, experts evaluated 3,200 anonymized answers from four systems, producing 12,800 criterion-level ratings.

\subsection*{Evidence retrieval and local revision}

OpenPerov Pro retrieved evidence from 173,024 article-centred documents through lexical, semantic and citation-based access. A source-retention selector chose 40 articles, a scientific-relevance reranker ordered them, and their evidence was assembled within a fixed budget \citep{Zhang2025Qwen3Embedding}. Pro identified scientific requirements left unresolved in the Flash answer and proposed evidence-linked additions or replacements. On PSM-Bench, Pro applied all structurally valid revisions (Figs.~\ref{fig:main2}c,d and \ref{fig:main3}b); critic-selected revision served as an alternative acceptance strategy. Both were compared with full-answer retrieval-augmented generation \citep{Lewis2020RAG} using the same answer-model checkpoint. SM6 describes retrieval, evidence assembly and revision.

\subsection*{Evaluation on subsequent studies and an external benchmark}

Ten studies published online after the literature-collection cutoff of 14 April 2026 supplied four questions per study in each of two settings: experimental context and limited context. The target articles and their supporting information were withheld. Flash answered without retrieval; Pro used the existing evidence collection and at most two edits per answer. An edit was accepted if it was scientifically relevant, consistent with the material and measurement regime, supported by sufficient evidence, and preserved the constraints of the question. Experts authored and checked their scientific answers using online resources available up to the same literature cutoff as OpenPerov. DeepSeek V4 Pro could assist with language editing. Independent volunteer perovskite experts evaluated the anonymized responses using a separate scientific rubric for each setting and finalized the scores through discussion. All comparisons retained study-level pairing (SM7 and Supplementary Table 1).

For the external benchmark, we used the 49 single-best-answer questions, reference options and model predictions released by the Perovskite-R1 authors \citep{Wang2026PerovskiteR1}. Flash answered locally with a fixed task-adapted prompt and without retrieval. Pro revised those answers using evidence from the top 40 articles, under an acceptance rule that permitted an option change when supported by retrieved evidence. Both routes used the same answer-model checkpoint. Predictions were scored by exact agreement with the released reference option (SM7.3).

\subsection*{SAM comparison and photovoltaic measurements}

The design task called for a carbazole contact at the ITO/perovskite interface. Archived prompts supplied the interfacial constraints and chemical evidence but withheld the experts' Na preference and the measured Na/K outcome. The resulting BCPCz--TMPS--Na and BCPCz--TMPS--K analogues were synthesized in-house and compared using the same solution preparation, SAM deposition and subsequent device fabrication, with 16 independent devices per group. Me-4PACz served as the conventional-SAM reference \citep{AlAshouri2020Extraction}. SM9 describes the design sequence.

SAM solutions in ethanol (0.5 mg mL$^{-1}$) were spin-coated onto cleaned ITO and annealed at 100 $^\circ$C. Small-area devices had the stack ITO/\allowbreak{}SAM/\allowbreak{}perovskite/\allowbreak{}PEACl/\allowbreak{}C$_{60}$/\allowbreak{}SnO$_{x}$/\allowbreak{}Ag, with MgF$_{2}$ on the glass side. Current-density--voltage measurements used a calibrated AM1.5G solar simulator at 100 mW cm$^{-2}$ and a 0.09 cm$^{2}$ aperture. Scans covered $-$0.1 to 1.22 V in both directions, with 20-mV steps and a 100-ms delay per step. External quantum efficiency was measured over 300--900 nm. Stabilized power output was recorded for approximately 600 s at fixed bias, with voltage magnitudes of 1.04 V for Me-4PACz and 1.06 V for SAM-K \citep{Christians2015Measurement}. SM10--SM12 detail the deposition and measurement procedures.

\subsection*{Stability protocols and statistical analysis}

Encapsulated devices were tested in ambient air under three ISOS-based protocols \citep{Khenkin2020ISOS}. Modified ISOS-LC-2 cycling comprised repeated 12-h illuminated periods at 85 $^\circ$C with MPP tracking, each followed by 12 h in the dark at approximately 20 $^\circ$C with tracking stopped. ISOS-D-2 testing used dark storage at 85 $^\circ$C with periodic photovoltaic measurements. ISOS-L-1 testing used continuous one-sun-equivalent illumination and MPP tracking under ambient-temperature conditions, with a device-surface temperature of approximately 45 $\pm$ 5 $^\circ$C. Cycling efficiencies were normalized to the first complete illuminated cycle. For continuous-MPP and dark-thermal tests, each group's efficiency was normalized to its initial value. Retentions were calculated from unsmoothed data, with rolling medians serving as visual guides (SM13).

Device populations are reported as arithmetic means and standard deviations. Stability time points represent repeated measurements of devices. Confidence intervals for model comparisons used 10,000 bootstrap resamples \citep{Efron1979Bootstrap}, preserving question-level pairing for PSM-Bench and study-level pairing for the subsequent-study tasks. SM14 provides the statistical definitions.

\section*{Data availability}
PSM-Bench, the subsequent-study expert-comparison tasks, reference answers and evaluation criteria, model responses, anonymized human assessments, and numerical evaluation records are publicly available at \url{https://github.com/Yan-Kai-Chen/OpenPerov}. The released component scores and analysis scripts reproduce the evaluation summaries reported here. Literature-derived training and retrieval corpora, source-support passages and evidence indexes are not publicly distributed.

\section*{Code availability}
Code for model adaptation, inference, evidence retrieval, local revision and evaluation is available at \url{https://github.com/Yan-Kai-Chen/OpenPerov}. The combined OpenPerov Flash adapter and the trained OpenPerov Pro retrieval adapters are available in the OpenPerov collection at \url{https://huggingface.co/collections/ChenYanKai2002/openperov-6abfd08c69d201a5e0533269}, together with loading instructions and model cards.

\section*{Author contributions}
Y.C. developed the large language model workflow, including corpus construction, model development and training, evidence retrieval and scientific evaluation. Z.W. conducted the perovskite experiments, materials characterization and device testing. Z.L. contributed to literature processing and corpus construction. T.J. supervised the research.

\section*{Competing interests}
The authors declare no competing interests.

\clearpage
\setcounter{page}{1}
\renewcommand{\thepage}{S\arabic{page}}
\setcounter{table}{0}
\renewcommand{\theHtable}{S\arabic{table}}
\onehalfspacing
\urlstyle{tt}
\renewcommand{\bibsection}{\section{Supplementary References}}
\begin{center}
  {\Large\bfseries Supplementary Information\par}
  \vspace{0.6em}
  {\LARGE\bfseries \ManuscriptTitle\par}
  \vspace{0.8em}
  {\large \AuthorLine\par}
  \vspace{0.6em}
  {\small\setstretch{1.1}\AffiliationBlock\par}
  \vspace{0.5em}
  {\footnotesize\setstretch{1.1}\AuthorNotes\par}
\end{center}
\vspace{0.6em}

\tableofcontents
\clearpage

\section{Supplementary Methods}
\subsection{SM1. Article-centred construction of the scientific corpus}

\subsubsection{SM1.1 Focal articles and source identity}
The OpenPerov corpus contains 4,532 curated articles on perovskite photovoltaics published in 2013--2026, including reviews and other relevant article types (Supplementary Fig.~S2). After reconciling article identities and versions, we assigned a stable CORE identifier to each article and its supporting information. The collection includes 25,848 figure-asset records and 3,819 table-asset records.

Derived passages, figure descriptions, table interpretations and syntheses inherited the training or holdout assignment of their parent article (SM2). Selected references and citing papers provided additional context and were counted separately from the curated CORE articles.

\subsubsection{SM1.2 Intrinsic evidence and literature lineage}
Perovskite-domain experts reconstructed the research logic of each study in an Integrated Article Evidence Record, linking its question and approach to experimental controls, observations and mechanistic interpretation under the relevant conditions. The main text, methods, supporting information, figures and tables supplied the evidence. Over several months, the experts calibrated every record by checking the scientific problem, its relationship to prior work, the experimental strategy and the evidence supporting the conclusions. GPT-5.5 and GPT-5.6 Sol assisted with text processing of the expert-constructed records. By comparison, related perovskite models generated passage-based question--answer and chain-of-thought examples (Perovskite-R1)~\citep{wang2026perovskiteR1} or extracted schema-defined knowledge and task-oriented answers (Perovskite-LLM and PVK-LLM)~\citep{liu2025perovskiteLLM,sun2026pvkLLM}.

Selected prior and subsequent studies supplied the literature lineage. Article matching, with the DOI as the primary identifier, resolved 36,625 distinct reference papers and 7,881 citing papers, which were linked to CORE articles through 101,779 and 12,406 deduplicated relations, respectively. Each relation retained its scientific role: earlier work established methods or motivated hypotheses, whereas later findings extended, qualified or contested interpretations. Supplementary Fig.~S4 shows the resulting citation communities.

\subsubsection{SM1.3 Figure and table evidence}
For Flash training, figure and table evidence was converted into scientific text linked to its source article. The descriptions distinguished measured observations, fitted quantities and proposed mechanisms. In the light-induced iodine-redistribution example (Supplementary Fig.~S5), the record connected the measured redistribution to the proposed role of mobile defects and identified the open questions about their identity and fate.

\subsection{SM2. Training corpora and article-level partitioning}

\subsubsection{SM2.1 Training and retrieval corpora}
The 4,035 adaptation articles supplied 78,299 structured text records covering intrinsic evidence, integrated article evidence, literature lineage and topic syntheses. The conventional comparator contained 211,518 PDF-derived instruction--answer records from the same articles: 105,760 closed-book and 105,758 context-grounded examples spanning mechanisms, stability, fabrication, performance, controls, figures, tables and scientific synthesis. The two routes used the same source articles and backbone. Each followed its own initial adaptation objective before receiving the same scientific instruction and answer-style training (SM3). The structured route produced OpenPerov Flash.

OpenPerov Pro retrieves from 173,024 documents spanning all 4,532 curated articles (SM6). These documents comprise article profiles, grouped evidence and individual statements. A separate derived collection contains 23,055 article-topic summaries. Article, training-record and retrieval-document counts therefore refer to different units.

\subsubsection{SM2.2 Article-level training boundary}
Let $\mathcal A$ denote the 4,532 curated articles and $\mathcal A_{\mathrm{train}}$ the 4,035 adaptation articles. The remaining 497 form $\mathcal A_{\mathrm{hold}}=\mathcal A\setminus\mathcal A_{\mathrm{train}}$, with $\mathcal A_{\mathrm{train}}\cap\mathcal A_{\mathrm{hold}}=\varnothing$. Every derived record inherited the assignment of its source article, so both adaptation corpora covered the same training articles and excluded held-out articles and benchmark answers. We also screened the adaptation data for overlap with the key scientific conclusions of the held-out studies and removed overlapping content. This screening covered both intrinsic article evidence and context from references or citing papers. Citation context was retained when it provided background without reproducing those held-out conclusions.

The unadapted base model provides the reference for assessing domain adaptation. Both learning routes start from the same pretrained backbone, holding the initial model state constant (SM3). At inference, Pro retrieves from the complete curated collection, including held-out articles. Closed-book systems receive only the question and task instructions (SM4, SM6).

\subsubsection{SM2.3 Preparation of training text}
Structured training text combined article evidence and scientific topic syntheses from the training partition. Records were normalized and deduplicated before training.

\subsection{SM3. Domain adaptation and scientific instruction training of OpenPerov Flash}

\subsubsection{SM3.1 Training sequence}
OpenPerov Flash was adapted from Qwen3.6-27B~\citep{qwen2026qwen36} through domain-adaptive pretraining (DAPT), targeted scientific instruction training and closed-book answer-style tuning. Each stage used low-rank adaptation (LoRA)~\citep{hu2022lora}, with the trained adapter merged into the backbone before the next stage.

\subsubsection{SM3.2 Domain-adaptive pretraining}
DAPT continued next-token pretraining on the 78,299 structured scientific records described in SM2~\citep{gururangan2020dont}. The corpus contained 19.33 million training tokens and linked research questions, experimental evidence and mechanistic interpretations.

\subsubsection{SM3.3 Targeted scientific training}
Scientific instruction training used 43,840 instruction--answer examples~\citep{wei2022flan}, followed by answer-style tuning with 7,490 training and 400 validation examples. Both stages used examples derived from the training articles and supervised only assistant-answer tokens. Scientific instruction training used a sample-weighted loss normalized by answer length; answer-style tuning used unweighted token-level cross-entropy.

\subsubsection{SM3.4 Conventional scientific corpus comparison and component ablations}
We compared two complete learning routes built from the same 4,035 source articles and the same Qwen3.6-27B backbone. For initial adaptation, the conventional route used assistant-only causal language modelling on 211,518 instruction--answer records (63.39 million training tokens); the structured route used text-only DAPT on 78,299 structured records (19.33 million training tokens; SM3.2). After their route-specific initial adaptation, the two routes scored 71.5 and 82.3, respectively, compared with 75.8 for the unadapted backbone (Fig.~2b). The 10.77-point difference is calculated from unrounded scores. Both checkpoints then received the same scientific instruction and answer-style training described in SM3.3.

After these shared training stages, the conventional and structured routes scored 84.07 and 87.57, respectively, a paired gain of 3.50 points (95\% paired-bootstrap interval~\citep{efron1979bootstrap}, 2.60--4.34; Supplementary Fig.~S12). Removing DAPT reduced the full 27B score to 85.98; omitting both the subsequent scientific-instruction and answer-style training stages left the DAPT-only checkpoint at 82.31. The unadapted model scored 75.83 (Supplementary Fig.~S13). Supplementary Fig.~S14 extends the comparison to 8B, 14B and 32B models: each scale includes the foundation and scientifically trained checkpoints, and the 27B and 32B scales also include a DAPT-only checkpoint.

\subsection{SM4. PSM-Bench construction and evaluation protocol}

\subsubsection{SM4.1 Scientific task definition and item construction}
PSM-Bench comprises 800 open-ended, closed-book questions grounded in 446 perovskite research articles. Each item contains a scientific problem, an expected line of reasoning, four scoring criteria, acceptable alternatives and common errors. Each question requires the interpretation of evidence and a reasoned explanation or experimental choice. Domain experts developed and finalized the content and criteria, and GPT-5.5 xhigh and GPT-5.6 Sol Pro assisted with checks of clarity, answerability and scoring consistency before the benchmark was frozen.

Its four task families are mechanism diagnosis ($n=368$), stability-constrained design ($n=300$), stability-failure analysis ($n=73$) and cross-constraint synthesis ($n=59$).

\subsubsection{SM4.2 Scientific reasoning requirements}
We annotated four requirements that could co-occur within a question and its scoring criteria: experimental discrimination (E), mechanistic explanation (M), design or decision logic (D), and alternatives or applicability boundaries (B). A deterministic text audit identified E in 746 items, M in 650, D in 599 and B in 469.

\subsubsection{SM4.3 Closed-book prompts and inference}
All 13 closed-book systems received the same frozen questions and a common system instruction requesting a final answer to a scientific examination question on perovskite photovoltaics. Task-adapted instructions specified the expected response for each task family: causal and competing explanations with decisive controls for mechanism diagnosis; interventions, failure boundaries and qualification tests for design; time-ordered mechanisms and discriminating measurements for failure analysis; and a design satisfying the stated constraints for synthesis. Model inputs excluded source articles, retrieval passages, reference answers, item-specific criteria, accepted equivalents, common errors and scoring outputs. Prompts requested the final answer without hidden reasoning or benchmark commentary~\citep{liang2023helm}.

The generation temperature was set to 0.2. One final answer was scored per question and system. Supplementary Table~\ref{tab:leaderboard-model-registry} lists model identifiers and access routes.

\subsubsection{SM4.4 Model comparison and prompt robustness}
Closed-book analyses compared Flash with 12 external or unadapted-backbone systems, and the overall leaderboard additionally includes evidence-conditioned Pro. Prompt-robustness comparisons used Flash, DeepSeek V4 Pro, DeepSeek V4 Flash and Qwen 3.6 27B Base under task-adapted, precision-focused and coverage-focused instructions on the same 800 questions.

\subsubsection{SM4.5 Empirical difficulty and discrimination}
Item difficulty was $D_i=100-\frac{1}{12}\sum_{m=1}^{12}s_{im}$, where $s_{im}$ is the Scientific Mastery score of non-OpenPerov closed-book comparator $m$ on item $i$. Higher $D_i$ values indicate lower comparator performance. To estimate discrimination, we used a frozen 13-system reference panel comprising the 12 non-OpenPerov closed-book comparators and an earlier Qwen3-32B-based OpenPerov model. For each item $i$, we ranked the systems by their mean scores on the other 799 questions. Discrimination was the difference between the median scores of the top four and bottom four systems on item $i$, with the middle five omitted. Family-level summaries used the frozen task assignments.

\subsubsection{SM4.6 Item-level comparisons}
For each of the 800 paired questions, Supplementary Fig.~S16 classifies Flash's score as higher than, equal to or lower than that of the comparator. Win-fraction confidence intervals use the 2.5th and 97.5th percentiles of 10,000 bootstrap resamples of paired item IDs.

\subsection{SM5. Scientific scoring and human expert validation}

\subsubsection{SM5.1 Criterion-level evaluation}
Each response was evaluated against a frozen reference record containing four scientific criteria, reference claims and accepted equivalents. Scientific Mastery summarizes four scoring components: semantic adequacy, rubric coverage, supported-claim precision and decision/control quality. These components were combined as described below.

\subsubsection{SM5.2 Item-score calculation}
Scientific Mastery combined semantic adequacy $S$, rubric coverage $C$ and decision/control quality $D$, each on a 0--100 scale, with supported-claim precision $P$ on a 0--1 scale. The additive base was $A=(S+C+D)/3$, and the gated score was
\begin{equation}
G=A\sqrt{\min(1,S/80)}\sqrt{\min(1,P/0.75)}.
\end{equation}
The missing-criterion cap $K$ was 100, 85, 70 or 50 when zero, one, two or at least three criteria were missing, respectively. The final item score was
\begin{equation}
s=\max\bigl\{\min(G,K),\ A-20\bigr\}.
\end{equation}
The $A-20$ floor limits any reduction from the additive base to 20 points and overrides the cap when the floor is higher. Item scores were then aggregated for each system.

\subsubsection{SM5.3 Robust aggregation across items}
The primary system score was the Huber location estimate~\citep{huber1964robust} of its 800 item scores. Starting from $\tilde s=\operatorname{median}_i(s_i)$, with fixed scale $\hat\sigma=\operatorname{median}_i|s_i-\tilde s|/0.67448975$ and $c=1.345$, we iterated to the fixed point
\begin{equation}
H_c=\frac{\sum_i v_i(H_c)s_i}{\sum_i v_i(H_c)},\qquad
v_i(\mu)=\min\!\left(1,\frac{c\hat\sigma}{|s_i-\mu|}\right),
\end{equation}
where $v_i(\mu)=1$ when $s_i=\mu$. This procedure downweights scores far from the estimated location while retaining every item; if $\hat\sigma=0$, the estimator returns the arithmetic mean. All systems used the same procedure. Supplementary Fig.~S8 reports the Huber score and arithmetic mean separately.

\subsubsection{SM5.4 Scoring execution}
Model-assisted calibration preceded evaluation. Item scores were calculated from the component scores and missing-criterion counts (SM5.2) and then aggregated with the Huber estimator (SM5.3). Each response record retained its component scores, missing-criterion count and final item score for system-level analysis.

\subsubsection{SM5.5 Double-blind human expert assessment}
In a double-blind assessment, experts from the School of Materials Science and Engineering, Tsinghua University, evaluated 3,200 anonymized final answers from four closed-book systems across all 800 questions. Faculty members and students provided 12,800 criterion-level ratings without access to system identities or automated scores. Each criterion was rated as not met (0), partially met (0.5) or met (1). For each answer, the human criterion score was 100 times the mean of its four ratings. These scores were then averaged arithmetically across answers.

\subsubsection{SM5.6 Agreement and ranking analyses}
Agreement between human and automated assessments was calculated across 16 system-by-task-family means (Pearson's $r=0.707$; Spearman's $\rho=0.729$). Ranking stability was assessed with 10,000 bootstrap resamples of the 800 question IDs, preserving the four-system pairing. Flash ranked first in 100\% of automated-score resamples and 99.84\% of human-expert resamples.

\subsection{SM6. Evidence-conditioned local revision in OpenPerov Pro}

\subsubsection{SM6.1 Evidence representation and task-specific retrieval}
OpenPerov Pro retrieves from 173,024 article-centred documents spanning all 4,532 curated articles (SM2). In Fig.~3a, \emph{Knowledge extraction} supplies source-linked evidence and \emph{Ranking Compression} orders relevant articles and assembles their content within an evidence budget. Each evidence statement retains the experimental conditions, scientific role, source scope and provenance of its claim, and normalization preserves recorded measurements and missing values.

For question $q$, we interleaved and deduplicated candidates from reciprocal-rank fusion~\citep{cormack2009rrf}, learned ranking, multivector retrieval, source recovery and citation-graph access. A source-retention selector reduced the candidate article set $\mathcal C_q$ to 40 articles, which a scientific-relevance reranker then ordered:
\begin{equation}
\mathcal T_q=\operatorname{Top}_{40}(\mathcal C_q;r_{\mathrm{src}}),\qquad
\pi_q=\operatorname{argsort}^{\downarrow}_{d\in\mathcal T_q}r_{\mathrm{rel}}(q,d).
\end{equation}
The selector and reranker were adapted from Qwen3-Reranker-0.6B and Qwen3-Reranker-8B, respectively~\citep{zhang2025qwen3embedding}. Reranking changed the order of the Top 40 articles but not their membership. Both models received the question, article title and evidence excerpt and were adapted with LoRA using an nDCG@10-weighted pairwise ranking objective~\citep{burges2006nonsmooth}. The selector used binary source labels, whereas the reranker used confidence-weighted graded relevance.

The source-retention selector was trained to distinguish each question's source article from hard negatives retrieved through semantic and earlier retrieval routes. Article-facet text was available for 7,486 of the 7,490 training questions. The 400 validation questions shared no source articles with the training questions. Source articles were included in the training and validation lists; benchmark pools consisted only of retrieved candidates. Checkpoints were selected by validation source recall at Top 40.

For $N$ evaluation questions and budget $K$, source retention was $N^{-1}\sum_i\mathbf{1}[d_i^{\star}\in\mathcal T_i^{(K)}]$, where $d_i^{\star}$ is the registered source article. The 800-question benchmark pool contained 537,643 query--article pairs and included the source for 793 questions. Across all 800 questions, the selector retained the source article in the Top 40 for 728 questions (91.00\%), compared with 662 (82.75\%) under the preceding ranking, a paired gain of 8.25 percentage points (95\% bootstrap interval, 5.50--11.12; Supplementary Fig.~S18a,c).

The scientific-relevance reranker was trained on 2,767 questions and 105,203 query--article pairs; validation used 200 questions and 3,621 pairs. Relevance grades ranged from 0 to 3, with confidence-weighted labels drawn from formal model-assisted judgments, earlier hard-example judgments and calibrated broad-pool soft targets. Training excluded candidate articles associated with the final benchmark and questions that exactly matched benchmark questions. Checkpoints were selected by validation nDCG@10.

Retrieval evaluation used expert-corrected judgments for fixed 40-paper pools, separate from the training labels. nDCG@10 measured graded ordering~\citep{jarvelin2002cumulated}; Relevant@10 measured the fraction of the first ten papers graded at least 2; and Grade-3 MRR measured the mean reciprocal rank of the first grade-3 paper across evaluation questions. The reranker scored 0.8739, 0.9616 and 0.9583 on these metrics, respectively. Fixed equal-weight fusion and the machine-learning baseline achieved nDCG@10 values of 0.8107 and 0.5744, respectively (Supplementary Fig.~S18b).

Evidence assembly followed the Top-40 ordering. A deterministic procedure selected article overviews and question-relevant topic summaries. It prioritized complete sentences by question-term coverage, task-family terms and quantitative or experimental detail, and then removed near-duplicates. Character budgets were 1,900 for ranks 1--8, 1,300 for ranks 9--20 and 800 for ranks 21--40. Where topic summaries were unavailable, article abstracts were used instead. Assembly preserved complete sentences and numerical claims, stored source identifiers separately and required no further model calls. Across 800 questions, 32,000 article positions supplied 54,176 distinct evidence blocks. A stratified review of 300 blocks assessed their scientific meaning, evidence role and scope (Supplementary Fig.~S17).

\subsubsection{SM6.2 Evidence-linked local answer revision}
Pro starts from the Flash answer $a_i^{(0)}$, identifies unresolved scientific requirements and proposes evidence-linked additions or replacements. On PSM-Bench, Pro applied all structurally valid edits; a selective-revision alternative used a consequence critic to accept a subset. Let $\mathcal P_i^{\mathrm{valid}}$ denote the valid edits and $\mathcal P_i^{\mathrm{sel}}$ the critic-selected subset. With acceptance strategy $\pi$, the outputs were
\begin{equation}
a_i(\pi=\mathrm{critic})=\operatorname{Dedup}\!\left[\operatorname{Apply}(a_i^{(0)},\mathcal P_i^{\mathrm{sel}})\right],\qquad
a_i(\pi=\mathrm{valid})=\operatorname{Dedup}\!\left[\operatorname{Apply}(a_i^{(0)},\mathcal P_i^{\mathrm{valid}})\right].
\end{equation}
The Pro setting supplied the results in Fig.~2c,d, Fig.~3b and Supplementary Fig.~S19. Both strategies retained text outside the edited spans and removed duplicate sentences. Revision used the draft and retrieved evidence, with reference answers, scoring criteria and item scores withheld.

In the mixed-halide example, a local edit linked the reduced photoluminescence redshift to carrier extraction while accounting for continued halide migration, AgI formation and contact hysteresis. The post-cutoff evaluation used task-specific acceptance criteria and at most two edits per answer (SM7).

\subsubsection{SM6.3 Controlled route comparison and uncertainty}
We compared closed-book Flash, full-answer retrieval-augmented generation (RAG)~\citep{lewis2020rag}, and the two local-revision settings on the same 800 questions, using the same answer-model checkpoint and Scientific Mastery scorer. Direct RAG generated a new answer from the question and evidence packet at a temperature of 0.2. The local routes instead edited the frozen Flash answer. For route $r$, the robust effect was
\begin{equation}
\Delta_r=H_{1.345}(s_{1r},\ldots,s_{800r})-H_{1.345}(s_{1\mathrm{F}},\ldots,s_{800\mathrm{F}}),
\end{equation}
where $s_{ir}$ is the item score and $H_{1.345}$ is the Huber estimator (SM5). We classified $\delta_{ir}=s_{ir}-s_{i\mathrm{F}}\geq5$ as a substantial gain and $\delta_{ir}\leq-5$ as a substantial loss. Paired 95\% confidence intervals used 10,000 question-ID bootstrap resamples, with all route scores for a question kept together. Direct RAG, selective revision and Pro increased the robust score by $+0.22$, $+1.23$ and $+1.49$ points, respectively. Substantial gains and losses occurred on 25.75\% and 24.38\% of items for direct RAG, and on 15.75\% and 0.50\% for Pro (Supplementary Fig.~S19).

\subsection{SM7. Post-cutoff scientific reasoning and external benchmark evaluation}

\subsubsection{SM7.1 Study selection, tasks and information access}
Ten perovskite studies published online after the literature-collection cutoff of 14 April 2026 supplied four questions per study in each of two settings (Supplementary Table~\ref{tab:s20-studies}). Study identifiers linked each question to its responses and scores. The 40 \emph{Experimental context} questions supplied observations for four tasks: mechanism diagnosis, decisive experiment, evidence update and scientific decision. The 40 \emph{Limited context} questions supplied a problem and sparse observations for mechanism diagnosis, competing mechanisms, evidence update and falsifiable prediction. The target conclusions, articles and supporting information were withheld.

Flash answered without retrieval, whereas Pro used the frozen evidence collection with at most two edits per answer. An edit was accepted if it was scientifically relevant, appropriate to the material and measurement regime, supported by sufficient evidence, and preserved the constraints of the question. Across both settings, 147 edits changed 76 of 80 answers. Reference answers and rubrics were withheld during revision. Perovskite-domain faculty members and students at the School of Materials Science and Engineering, Tsinghua University, authored and checked the expert answers using online resources available up to the same literature cutoff as OpenPerov.

\subsubsection{SM7.2 Setting-specific scoring and study-paired analysis}
Independent volunteer perovskite experts evaluated the anonymized responses using a separate scientific rubric for each setting and finalized the scores through discussion. Experimental-context answers were evaluated against question-specific scientific criteria. Limited-context criteria covered mechanistic validity, use of sparse evidence, competing-mechanism discrimination, measurement--mechanism mapping, uncertainty calibration and falsifiability. These setting-specific scores are distinct from Scientific Mastery (SM5).

The reported Pro score equals the Flash score plus the mean paired Pro--Flash difference from the same scoring batch. Confidence intervals used 10,000 bootstrap resamples of the ten study identifiers, retaining all matched questions within each draw. The two information settings and their rubrics were analysed separately (Supplementary Fig.~S20).

\subsubsection{SM7.3 Independent Perovskite-R1 multiple-choice evaluation}
We evaluated transfer on the 49 single-best-answer questions in Perovskite-R1 Table S4~\citep{wang2026perovskiteR1}. Flash used a frozen task-adapted prompt without retrieval. Pro revised the Flash answers using a frozen Top-40 evidence ordering, under an acceptance rule that permitted an option change when supported by retrieved evidence. Both routes used the 27B OpenPerov backbone.

Exact-match scoring used the released reference options. Perovskite-R1 predictions came from the authors' release, and OpenPerov predictions were generated locally. Perovskite-R1, Flash and Pro answered 45/49, 46/49 and 48/49 questions correctly, respectively (Supplementary Fig.~S23).

\subsection{SM8. Research themes, citation networks and forward signals}

\subsubsection{SM8.1 Corpus themes and the CORE citation network}
We mapped the evidence records of 4,532 CORE papers to eight predefined themes: materials and composition, device architecture, fabrication and processing, interface engineering, defects and recombination, characterization and metrics, performance enhancement, and stability and degradation. For paper $i$, the support fraction $f_{it}$ was the proportion of its grouped evidence records carrying scientific-domain tag $t$. For theme $k$, let $\mathcal T_k$ denote its tag set and $\bar f_{ik}^{(3)}$ the mean of the three largest available tag fractions for paper $i$. We calculated
\begin{equation}
s_{ik}=0.78\max_{t\in\mathcal T_k}f_{it}+0.22\bar f_{ik}^{(3)},
\end{equation}
with $s_{ik}=0$ when no tag mapped to the theme. Scores were standardized within each paper using that paper's sample standard deviation. We retained up to four themes with $z_{ik}\geq0.50$, or the highest-scoring theme if none met this threshold, so assignments could overlap (Supplementary Fig.~S3a).

Predefined term patterns assigned eight subtopics to the 1,381 stability-related papers: operational lifetime, moisture/oxygen, thermal stress, light/UV stress, ion migration, phase instability, interfacial failure and encapsulation/barrier (Supplementary Fig.~S3b). The annual heat map reports theme counts normalized by each year's CORE article count for 2018--2025. The sparse 2013--2017 and incomplete 2026 records contribute to corpus-wide counts but fall outside this temporal display.

We matched citations first by normalized DOI and then by normalized title when the title match was unique. The network contained 4,532 CORE nodes and 26,402 directed citing--cited pairs. Community detection used the symmetrized topology of the largest weakly connected component (3,758 papers), with no theme, journal or year inputs. The remainder comprised 765 isolated papers and nine papers in smaller components. Leiden clustering~\citep{traag2019leiden} at resolution 1.0, which gave the highest mean modularity among the resolutions tested, yielded 11 communities. The displayed seed-4 partition showed the highest average agreement with the other runs at that resolution. Community--theme enrichment was
\begin{equation}
E_{ck}=\log_2\!\left[\frac{n_{ck}/n_c}{n_k/N}\right],\qquad N=4{,}532,
\end{equation}
where $n_{ck}$ counts papers in community $c$ assigned to theme $k$, $n_c$ is community size and $n_k$ is the corpus-wide theme count. Fisher's exact tests compared each community--theme pair with the rest of the corpus, with Benjamini--Hochberg correction~\citep{benjamini1995fdr}; Supplementary Fig.~S4c marks $q<0.05$. The broader literature lineage in Fig.~S4a was assembled separately from this CORE--CORE partition.

\subsubsection{SM8.2 Evidence-derived stability research landscape}
We constructed a stability research landscape using language-model extraction and concept graphs~\citep{marwitz2026materials}. DeepSeek V4 Flash extracted 5,251 distinct concepts from 1,381 stability CORE papers. Synonym merges were adjudicated to preserve concept class, numerical content, comparison scope and scientific conditions, whereas parent and child concepts, and related but distinct concepts, retained separate identities. The resulting 5,243 canonical groups retained links to their source occurrences.

Concepts were embedded with Qwen3-Embedding-0.6B~\citep{zhang2025qwen3embedding} and combined into normalized, occurrence-weighted group vectors. Graph edges represented four relationships: semantic similarity among the 12 nearest neighbours with cosine similarity of at least 0.58; co-occurrence within a paper; accepted parent--child relations; and adjudicated relations between distinct concepts. Edge weights were
\begin{equation}
w_{ab}=\max\!\left(0,\frac{c_{ab}-0.5}{0.5}\right)+0.20\log(1+m_{ab})+h_{ab}+0.10r_{ab},
\end{equation}
where $c_{ab}$ is the retained semantic similarity (zero when no semantic edge exists), $m_{ab}$ counts papers containing both groups, and $h_{ab}$ and $r_{ab}$ indicate parent--child and related-but-distinct relations, respectively. The graph contained 52,932 edges. Louvain clustering~\citep{blondel2008fast} at resolution 2.0 and seed 17 produced 32 macro themes, with partition selection guided by cross-seed agreement, modularity and cluster-size diagnostics. Experts reviewed and confirmed the model-assisted theme labels. Supplementary Fig.~S21 presents the all-paper map; for the historical analyses, themes were rebuilt at each cutoff.

For theme $f$ and interval $T$, article-normalized prevalence was
\begin{equation}
P_f(T)=\frac{\bigl|\{i\in\mathcal C_T:i\text{ contains theme }f\}\bigr|}{|\mathcal C_T|},
\end{equation}
where $\mathcal C_T$ contains the stability CORE papers published in $T$. Each paper counted once per assigned theme and could belong to several themes. Supplementary Fig.~S21b covers 2013--2018 ($n=116$), 2019--2021 ($n=161$) and 2022--2025 ($n=1{,}029$). These complete-period denominators exclude partial-2026 papers and five papers with unresolved publication years; the latter remain in the concept registry.

\subsubsection{SM8.3 Historical and partial-2026 forward-signal analyses}
We tested whether signals derived from the landscape could identify themes that later attracted more research attention. Graphs rebuilt at the 2018, 2021 and 2023 cutoffs contained 38, 44 and 52 themes, respectively. The corresponding baseline/recent/future windows were 2013--2015/2016--2018/2019--2021, 2016--2018/2019--2021/2022--2023, and 2019--2021/2022--2023/2024--2025, respectively. Dated citation trajectories used OpenAlex-indexed citing works~\citep{priem2022openalex}. Downstream uptake was calculated from direct references between stability CORE papers matched through OpenAlex identifiers. With $\mathcal H_f$ denoting papers assigned to theme $f$ by the cutoff and $\mathcal R_i$ the references of paper $i$,
\begin{equation}
U_f(T)=\frac{1}{|\mathcal C_T|}\sum_{i\in\mathcal C_T}\mathbf{1}\!\left[\mathcal R_i\cap\mathcal H_f\neq\varnothing\right],\qquad
\Delta U_f=U_f(T_{\mathrm{future}})-U_f(T_{\mathrm{recent}}).
\end{equation}
Uptake is therefore the fraction of stability papers in $\mathcal C_T$ that cite a theme's pre-cutoff literature.

We calculated two pre-cutoff signals. Citation acceleration $A_f$ compared smoothed citation rates in recent ($T_r$) and baseline ($T_b$) windows. $C_f(T)$ counts dated OpenAlex citations to linked historical theme papers, and $E_f(T)$ sums their observable paper-years, starting at publication or the start of $T$, whichever is later. Novel recombination $R_f$ measured the fraction of recent within-paper theme-pair events involving $f$ that had not occurred before $T_r$; it was zero when no pair event occurred. The combined signal was
\begin{align}
A_f&=\log\frac{C_f(T_r)+0.5}{E_f(T_r)+1}-\log\frac{C_f(T_b)+0.5}{E_f(T_b)+1},\\
Z_f&=\tfrac12\left[z(A_f)+z(R_f)\right].
\end{align}
where $z(\cdot)$ denotes mean-centred standardization by the sample standard deviation across themes at that cutoff. We calculated Spearman correlations of each signal with $\Delta U_f$ and used recent uptake as the popularity baseline (Supplementary Fig.~S22a).

A graph built through 2025 ranked 48 themes for near-term monitoring. We compared partial-2026 uptake changes in the top-five and top-ten groups with the remaining themes using one-sided permutation tests with 10,000 label permutations. Across all 48 themes, we calculated adjusted Spearman associations from average ranks after residualizing them against article count and 2024--2025 uptake. The 95\% confidence intervals used 10,000 paper-level bootstrap resamples of the 70 partial-2026 papers, holding the 2025 themes, predictors and ranking fixed (Supplementary Fig.~S22b).

\subsection{SM9. Human--model design of a carbazole self-assembled monolayer}

\subsubsection{SM9.1 Design task and information sequence}
The design task targeted a solution-processable carbazole contact at the hydroxylated indium tin oxide (ITO)/self-assembled monolayer (SAM)/perovskite interface. The brief required oxide anchoring and monolayer integrity, together with control of undercoordinated Pb and interfacial ions. It called for an alternative to phosphonic- or phosphoric-acid anchoring, with passivation functionality built into the SAM itself.

In round A, Flash compared up to three approaches, assigned functions to the carbazole N and 3,6 positions, identified possible failure modes and proposed falsification tests. The brief left both the molecular architecture and the choice of alkali metal open.

Round B retained the question and the frozen Flash answer and added two chemical evidence statements: trialkoxysilanes offer oxide-facing silanol chemistry with a risk of excessive condensation, and carboxylates can interact with undercoordinated Pb while their alkali-metal forms could alter the ionic environment. Trimethoxysilane anchoring has precedent in SAM hole-transport layers for inverted perovskite cells~\citep{tang2024sam}. The model was asked to assign functions to the N and 3,6 positions, compare one versus two carboxylate arms and methoxy versus ethoxy silanes, and choose Li, Na or K. Round C retained the earlier prompts and the frozen Flash draft, and requested explicit molecular connectivity together with experiments to test the proposed mechanisms. Pro applied evidence-conditioned local revision in rounds B and C (SM6). The experts' Na preference and the measured Na/K outcome were withheld throughout.

\subsubsection{SM9.2 Architecture and counterion choice}
In round B, the proposed architecture placed an ITO-facing trimethoxysilyl branch at the carbazole N position and two perovskite-facing carboxylate branches at positions 3 and 6. Flash selected K as the first counterion to test. Pro retained K and specified a carboxylic-acid or no-salt control, matched Li/Na/K analogues, and tests of anchoring, Pb coordination, ion redistribution and processing stability.

Round C specified the connectivity of BCPCz--TMPS--K: a para-phenylene linker from carbazole N9 to a trimethoxysilyl group and para-phenylene linkers from C3 and C6 to potassium carboxylates. The proposed experiments were designed to distinguish ITO attachment from lateral siloxane condensation and to test carboxylate--Pb coordination, K redistribution, interface energetics and charge extraction. The model also identified excessive crosslinking and counterion-dependent crystallization or migration as possible failure modes.

\subsubsection{SM9.3 Expert--model choice and experimental comparison}
Figure~4a,b summarizes the design exchange and the contrast between the experts' Na preference and OpenPerov's K choice. Devices incorporating the K analogue achieved a higher mean efficiency than those incorporating the Na analogue (Supplementary Fig.~S24 and Supplementary Table~2), supporting the model's counterion selection. Archived prompts and responses document the sequence described in SM9.1--SM9.2. The subsequent comparison with conventional Me-4PACz uses BCPCz--TMPS--K as Target and Me-4PACz as Control (Fig.~4c--h); fabrication and measurements are described in SM10--SM13.

\subsection{SM10. Materials and self-assembled monolayer solutions}

BCPCz--TMPS--Na and BCPCz--TMPS--K were synthesized in-house. Me-4PACz served as the reference SAM~\citep{alashouri2020me4pacz}. The device materials were patterned ITO substrates, PbI$_2$, formamidinium iodide (FAI), methylammonium chloride (MACl), PbCl$_2$, phenethylammonium chloride (PEACl), C$_{60}$, SnO$_x$ and Ag. Solvents were N,N-dimethylformamide (DMF), dimethyl sulfoxide (DMSO), chlorobenzene, ethanol and isopropanol.

Each SAM was dissolved separately in ethanol at 0.5~mg~mL$^{-1}$ and sonicated for 10~min. The same solution-preparation and deposition procedures were used for all three SAMs.

\subsection{SM11. Perovskite solar-cell fabrication}

\subsubsection{SM11.1 Device architecture and substrate preparation}
Small-area cells had the stack ITO/SAM/perovskite/PEACl/C$_{60}$/SnO$_x$/Ag with a glass-side MgF$_2$ antireflection layer. The module schematic in Fig.~4a uses Cu and P1/P2/P3 laser patterning. After SAM deposition, the same fabrication procedure was used for small-area cells incorporating Me-4PACz, BCPCz--TMPS--Na and BCPCz--TMPS--K. Patterned ITO substrates (12~$\Omega$~sq$^{-1}$) were washed with detergent and then sonicated for 15~min each in deionized water, acetone, absolute ethanol and isopropanol. The substrates were dried under N$_2$ and treated with ultraviolet--ozone for 20~min.

\subsubsection{SM11.2 SAM and perovskite deposition}
SAM solutions (SM10) were spin-coated onto cleaned ITO at 3,000~rpm for 30~s, annealed at 100~$^\circ$C for 15~min in an N$_2$ glovebox and cooled before perovskite deposition.

A 1.5~M FAPbI$_3$-based precursor solution was prepared from PbI$_2$ (839.0~mg), FAI (309.6~mg), MACl (18.9~mg) and PbCl$_2$ (12.7~mg) in DMF/DMSO (8:2, v/v). The solution was stirred at 60~$^\circ$C for 2~h in an N$_2$ glovebox and passed through a 0.22-$\mu$m polytetrafluoroethylene filter. The precursor solution was spin-coated at 1,000~rpm for 10~s and then 5,000~rpm for 40~s. Chlorobenzene (150~$\mu$L) was dispensed 5~s before the end of the second step. The perovskite films were annealed at 100~$^\circ$C for 45~min, coated with a 1.0~mg~mL$^{-1}$ PEACl solution in isopropanol and annealed again at 100~$^\circ$C for 5~min.

\subsubsection{SM11.3 Electron-selective contact and electrode}
The electron-selective contact and electrode comprised approximately 20~nm C$_{60}$, 20~nm SnO$_x$ and 120~nm Ag, deposited in that order. C$_{60}$ and Ag were thermally evaporated at approximately $4.0\times10^{-4}$~Pa, and SnO$_x$ was deposited by atomic layer deposition. A 100-nm MgF$_2$ antireflection layer was then deposited on the glass side. SM12 specifies the measurement aperture.

\subsection{SM12. Photovoltaic characterization}

\subsubsection{SM12.1 Current-density--voltage measurements}
Photovoltaic performance was measured at room temperature in an N$_2$ glovebox using a Keithley 2400-series source meter and a Newport 94063A Oriel Sol3A Class AAA simulator. Illumination was calibrated to AM1.5G at 100~mW~cm$^{-2}$ using a Si reference cell and a KG-5 filter. A metal aperture mask with a 0.09~cm$^2$ opening defined the illuminated measurement area. Current-density--voltage ($J$--$V$) scans covered $-0.1$ to 1.22~V in both directions to capture scan-dependent hysteresis~\citep{snaith2014hysteresis}, using 20-mV steps and a 100-ms delay per step (200~mV~s$^{-1}$). Open-circuit voltage ($V_\mathrm{OC}$), short-circuit current density ($J_\mathrm{SC}$), fill factor and power conversion efficiency were obtained from these curves.

\subsubsection{SM12.2 External quantum efficiency and stabilized output}
External quantum efficiency (EQE) was measured at zero applied bias using an Enli Tech QE-R 3011 system. Spectra spanned 300--900~nm in 10-nm steps (Fig.~4d), and $J_\mathrm{SC}$ was obtained by integration against the AM1.5G reference spectrum. Stabilized power output was recorded for approximately 600~s at fixed voltage (Fig.~4e)~\citep{christians2015measurements}, using voltage magnitudes of 1.04~V for Control and 1.06~V for Target. SM13 describes long-duration maximum-power-point (MPP) tracking.

\subsubsection{SM12.3 Device populations}
Figure~4c--h compares Me-4PACz (Control) with BCPCz--TMPS--K (Target); panels c--e show representative traces. The counterion comparison uses independent groups of 16 BCPCz--TMPS--Na and 16 BCPCz--TMPS--K devices (Supplementary Fig.~S24 and Supplementary Table~2). The individual device was the sampling unit, and SM14 defines the population statistics.

\subsection{SM13. Stability testing and time-series processing}

Encapsulated cells were tested in ambient air under all three stability protocols.

\subsubsection{SM13.1 Light--dark cycling}
Light--dark cycling followed a modified ISOS-LC-2 protocol~\citep{khenkin2020isos} (Fig.~4f). Each 24-h cycle comprised 12~h at $85\,^{\circ}\mathrm{C}$ under 100~mW~cm$^{-2}$ white-LED illumination with maximum-power-point (MPP) tracking on a PURI2400-E8 system, followed by 12~h in the dark at approximately $20\,^{\circ}\mathrm{C}$ with tracking stopped.

\subsubsection{SM13.2 Continuous maximum-power-point operation}
Continuous MPP testing used a PURI2400-E8 system under one-sun-equivalent white-LED illumination and ambient-temperature ISOS-L-1 conditions~\citep{khenkin2020isos} (Fig.~4h). The device-surface temperature under illumination was approximately $45\pm5\,^{\circ}\mathrm{C}$.

\subsubsection{SM13.3 Dark thermal ageing}
Devices were stored in the dark at $85\,^{\circ}\mathrm{C}$ following the ISOS-D-2 protocol~\citep{khenkin2020isos}, with periodic photovoltaic measurements (Fig.~4g).

\subsubsection{SM13.4 Processing and normalization of stability traces}
Cycling efficiencies were averaged over complete 12-h illuminated segments, excluding the incomplete final segment, and normalized to the first complete light-on cycle. Target cycle means were calculated from instrument-exported time series; Control cycle means were extracted by digitizing the corresponding experimental trace.

For continuous MPP and dark thermal ageing, each group's efficiency was normalized to its initial value:
\begin{equation}
  R(t)=\frac{\mathrm{PCE}(t)}{\mathrm{PCE}(0)}.
\end{equation}
The initial absolute PCEs in the MPP test were 24.32\% for Control and 24.99\% for Target. Both MPP records covered 994~h of elapsed experimental time. In the thermal test, Target retained 96.74\% of its initial efficiency after 1,000~h, and the Control record ended at 730~h. Pale traces show measured values, and centred rolling medians over 15 points (thermal) and 25 points (MPP) guide the eye. All reported retentions were calculated from unsmoothed data.

\subsection{SM14. Statistical analysis and reporting}

\subsubsection{SM14.1 Units of analysis and uncertainty}
Analyses used frozen question, response and score records. Paired PSM-Bench bootstrap analyses resampled questions (SM4--SM6); post-cutoff analyses resampled studies while retaining matched questions (SM7). Temporal-theme analyses used paper-level units (SM8). Each bootstrap analysis used 10,000 resamples, and the resulting intervals are reported alongside the corresponding estimates.

\subsubsection{SM14.2 Device statistics}
The Na and K device groups were independent and are summarized by arithmetic means and standard deviations in Supplementary Table~\ref{tab:s24-photovoltaic-parameters}. Supplementary Fig.~S24 shows individual devices, medians, interquartile ranges and whiskers extending to the most extreme observations within 1.5 times the interquartile range. Stability traces comprise repeated measurements of devices and were processed and normalized as described in SM13.

\subsubsection{SM14.3 Use of large language models and author responsibility}
Domain experts constructed and calibrated the scientific reasoning in the corpus, and GPT-5.5 and GPT-5.6 Sol assisted with text processing. LLMs also supported benchmark calibration and molecular design and were themselves evaluated as research systems (SM1--SM9). General-purpose LLM tools assisted with code debugging and manuscript editing. The authors designed the research, conducted the experiments, evaluated model-assisted suggestions and interpreted the results. They take full responsibility for the accuracy and integrity of the work.

\clearpage
\section{Supplementary Tables}

\renewcommand{\tablename}{Supplementary Table}

\begin{longtable}{@{}
  >{\raggedright\arraybackslash}p{1.05cm}
  >{\raggedright\arraybackslash}p{6.10cm}
  >{\raggedright\arraybackslash}p{1.95cm}
  >{\raggedright\arraybackslash}p{1.95cm}
  >{\raggedright\arraybackslash}p{3.20cm}@{}}
\caption{Studies used in the two post-cutoff scientific-reasoning settings (Supplementary Fig.~S20).}
\label{tab:s20-studies}
\addcontentsline{toc}{subsection}{Table 1. Studies used in the two post-cutoff scientific-reasoning settings (Supplementary Fig.~S20)}\\
\toprule
Study ID & Article & Journal & Online publication & DOI \\
\midrule
\endfirsthead

\multicolumn{5}{l}{\footnotesize\textit{Supplementary Table 1 continued}}\\
\toprule
Study ID & Article & Journal & Online publication & DOI \\
\midrule
\endhead

\midrule
\multicolumn{5}{r}{\footnotesize Continued on next page}\\
\endfoot

\bottomrule
\endlastfoot

S20-EC01 & Bypassing the yellow phase for extremely stable formamidinium lead iodide perovskite solar cells & Science & 30 April 2026 & \href{https://doi.org/10.1126/science.aeb7992}{\nolinkurl{10.1126/science.aeb7992}} \\

S20-EC02 & Stereoelectronic manipulation of ligands for perovskite solar cells & Nature & 13 May 2026 & \href{https://doi.org/10.1038/s41586-026-10626-0}{\nolinkurl{10.1038/s41586-026-10626-0}} \\

S20-EC03 & Omnidirectional ionic locking network for stable perovskite photovoltaics & Nature Photonics & 29 May 2026 & \href{https://doi.org/10.1038/s41566-026-01918-y}{\nolinkurl{10.1038/s41566-026-01918-y}} \\

S20-EC04 & Perovskite--organic tandem solar cells with a photo-transformable stabilizer & Nature & 13 July 2026 & \href{https://doi.org/10.1038/s41586-026-10869-x}{\nolinkurl{10.1038/s41586-026-10869-x}} \\

S20-EC05 & Controlled solid-state crystallization with formamidinium acetate for fully vacuum-deposited perovskite solar cells & Nature Energy & 14 July 2026 & \href{https://doi.org/10.1038/s41560-026-02093-8}{\nolinkurl{10.1038/s41560-026-02093-8}} \\

S20-EC06 & An entropy-regulating molecular lock stabilizes formamidinium lead halide perovskite & Science & 14 May 2026 & \href{https://doi.org/10.1126/science.aeb9953}{\nolinkurl{10.1126/science.aeb9953}} \\

S20-EC07 & Spatiotemporally homogeneous crystallization for ambient scalable perovskite photovoltaics & Science & 16 July 2026 & \href{https://doi.org/10.1126/science.aef1969}{\nolinkurl{10.1126/science.aef1969}} \\

S20-EC08 & Close-space sublimation as a versatile deposition process for efficient perovskite silicon tandem solar cells & Nature Energy & 19 May 2026 & \href{https://doi.org/10.1038/s41560-026-02068-9}{\nolinkurl{10.1038/s41560-026-02068-9}} \\

S20-EC09 & Electronic-resonance enhanced molecule for perovskite solar cells & Nature & 22 July 2026 & \href{https://doi.org/10.1038/s41586-026-10919-4}{\nolinkurl{10.1038/s41586-026-10919-4}} \\

S20-EC10 & Electrodeposited self-assembled molecules for perovskite photovoltaics & Nature & 2 July 2026 & \href{https://doi.org/10.1038/s41586-026-10844-6}{\nolinkurl{10.1038/s41586-026-10844-6}} \\

\end{longtable}

\noindent\footnotesize Each study contributed four questions in each of the two settings, Experimental context and Limited context (SM7; Supplementary Fig.~S20). Dates are online publication dates; all are after the literature-collection cutoff of 14 April 2026. Each setting used its own information conditions and scoring rubric.\normalsize

\begin{table}[htbp]
\centering
\caption{Photovoltaic parameters of BCPCz--TMPS--Na and BCPCz--TMPS--K devices.}
\label{tab:s24-photovoltaic-parameters}
\addcontentsline{toc}{subsection}{Table 2. Photovoltaic parameters of BCPCz--TMPS--Na/K devices}
\begin{tabular}{@{}lccccc@{}}
\toprule
SAM & $n$ & $V_\mathrm{OC}$ (V) & $J_\mathrm{SC}$ (mA cm$^{-2}$) & Fill factor (\%) & PCE (\%) \\
\midrule
BCPCz--TMPS--Na & 16 & $1.174 \pm 0.003$ & $25.82 \pm 0.24$ & $83.28 \pm 0.75$ & $25.24 \pm 0.30$ \\
BCPCz--TMPS--K  & 16 & $1.191 \pm 0.003$ & $26.15 \pm 0.18$ & $85.71 \pm 0.48$ & $26.69 \pm 0.27$ \\
\bottomrule
\end{tabular}

\vspace{1.5mm}
\begin{minipage}{0.96\textwidth}
\footnotesize Values are reported as mean $\pm$ standard deviation for independent groups of 16 devices per SAM. All 32 measurements appear in Supplementary Fig.~S24.
\end{minipage}
\end{table}

\clearpage
\begin{landscape}
\begingroup
\scriptsize
\setlength{\tabcolsep}{2pt}
\renewcommand{\arraystretch}{1.18}
\setlength{\LTleft}{0pt}
\setlength{\LTright}{0pt}

\captionof{table}{Model and system configurations in the Figure 2 leaderboard.}
\label{tab:leaderboard-model-registry}
\addcontentsline{toc}{subsection}{Table 3. Model and system configurations in the Figure 2 leaderboard}
\vspace{1mm}

\begin{longtable}{@{}
  >{\centering\arraybackslash}p{0.80cm}
  >{\raggedright\arraybackslash}p{2.25cm}
  >{\raggedright\arraybackslash}p{4.20cm}
  >{\raggedright\arraybackslash}p{3.55cm}
  >{\raggedright\arraybackslash}p{1.75cm}
  >{\raggedright\arraybackslash}p{2.75cm}
  >{\raggedright\arraybackslash}p{2.45cm}
  >{\raggedright\arraybackslash}p{2.70cm}@{}}
\toprule
Order & Leaderboard label & Exact evaluated model or system & Access in this study & Evaluation mode & Architecture & Total parameters & Activated parameters per token \\
\midrule
\endfirsthead

\multicolumn{8}{l}{\footnotesize\textit{Supplementary Table 3 continued}}\\
\toprule
Order & Leaderboard label & Exact evaluated model or system & Access in this study & Evaluation mode & Architecture & Total parameters & Activated parameters per token \\
\midrule
\endhead

\midrule
\multicolumn{8}{r}{\footnotesize Continued on next page}\\
\endfoot

\bottomrule
\endlastfoot

1 & OpenPerov Pro & OpenPerov Flash backbone with evidence retrieval and local revision & Local composite system & Evidence-conditioned & Dense 27B backbone plus retrieval and revision pipeline & 27B backbone & 27B backbone \\

2 & OpenPerov Flash & OpenPerov Flash, adapted from Qwen3.6-27B & Local weights & Closed-book & Dense & 27B language-model backbone & All backbone parameters (27B) \\

3 & GLM-5.2 & API identifier: \texttt{GLM-5.2} & Paratera OpenAI-compatible API; evaluated 24 July 2026 & Closed-book & MoE & $\sim$743B\textsuperscript{a} & $\sim$39B\textsuperscript{a} \\

4 & Kimi-K2.6 & API identifier: \texttt{Kimi-K2.6} & Paratera OpenAI-compatible API; evaluated 24 July 2026 & Closed-book & MoE & 1T & 32B \\

5 & Qwen 3.7 Max & Provider identifier: \texttt{Qwen3.7-Max} & Paratera OpenAI-compatible API; evaluated 25 July 2026 & Closed-book & Not publicly disclosed & Not publicly disclosed & Not publicly disclosed \\

6 & Gemini 3.1 Pro & Provider alias: \nolinkurl{gemini-3.1-pro-preview-maxthinking}; official family: \nolinkurl{gemini-3.1-pro-preview} & Paratera OpenAI-compatible API; evaluated 25 July 2026 & Closed-book & Not publicly disclosed & Not publicly disclosed & Not publicly disclosed \\

7 & DeepSeek V4 Pro & API identifier: \texttt{DeepSeek-V4-Pro} & Paratera OpenAI-compatible API; evaluated 21 July 2026 & Closed-book & MoE & 1.6T & 49B \\

8 & Gemini 3.5 Flash & Provider alias: \nolinkurl{gemini-3.5-flash-maxthinking}; official family: \nolinkurl{gemini-3.5-flash} & Paratera OpenAI-compatible API; evaluated 25 July 2026 & Closed-book & Not publicly disclosed & Not publicly disclosed & Not publicly disclosed \\

9 & DeepSeek V4 Flash & API identifier: \texttt{DeepSeek-V4-Flash-0731} & Paratera OpenAI-compatible API; evaluated 11 August 2026 & Closed-book & MoE & 285B & 13B \\

10 & Qwen 3.6 27B Base & Local checkpoint corresponding to \nolinkurl{Qwen/Qwen3.6-27B} & Local weights & Closed-book & Dense & 27B language model\textsuperscript{b} & All language-model parameters (27B) \\

11 & Qwen3 32B & \nolinkurl{Qwen/Qwen3-32B} & Local weights & Closed-book & Dense & 32.8B total; 31.2B non-embedding & All parameters (32.8B) \\

12 & Gemma-4-31B & \nolinkurl{google/gemma-4-31B-it} & Local weights & Closed-book & Dense & $\sim$30.7B\textsuperscript{c} & All language-model parameters ($\sim$30.7B) \\

13 & DeepSeek R1 & API identifier: \texttt{DeepSeek-R1} & Paratera OpenAI-compatible API; evaluated 24 July 2026 & Closed-book & MoE & 671B & 37B \\

14 & Perovskite-R1 & \nolinkurl{JH976/Perovskite-R1}; official merged BF16 checkpoint & Local weights & Closed-book & Dense; LoRA-fine-tuned QwQ-32B & $\sim$32.5B total; $\sim$31B non-embedding & All parameters ($\sim$32.5B) \\

\end{longtable}

\vspace{-2mm}
\noindent\textsuperscript{a}GLM-5.2 counts follow the architecture-level estimate of approximately 743B total and 39B activated parameters; packaged-model metadata may report 753B total. \textsuperscript{b}``Base'' denotes the unadapted OpenPerov reference checkpoint; the upstream Qwen3.6-27B release is post-trained. \textsuperscript{c}Gemma counts refer to the language-model component; its vision encoder was unused.

\smallskip
\noindent Pro uses the Flash 27B backbone with retrieval and local revision. Perovskite-R1 was LoRA-fine-tuned from QwQ-32B, following the lineage Qwen2.5-32B $\rightarrow$ QwQ-32B $\rightarrow$ Perovskite-R1. API identifiers and evaluation dates specify the versions evaluated.

\smallskip
\noindent\textit{Architecture sources:} \href{https://huggingface.co/zai-org/GLM-5.2}{GLM-5.2}; \href{https://github.com/vllm-project/recipes/blob/main/models/zai-org/GLM-5.2.yaml}{GLM-5.2 runtime recipe}; \href{https://huggingface.co/moonshotai/Kimi-K2.6}{Kimi-K2.6}; \href{https://help.aliyun.com/en/model-studio/qwen3-7-max}{Qwen3.7-Max}; \href{https://ai.google.dev/gemini-api/docs/models}{Gemini}; \href{https://fe-static.deepseek.com/chat/transparency/deepseek-V4-model-card-EN.pdf}{DeepSeek V4}; \href{https://huggingface.co/Qwen/Qwen3.6-27B}{Qwen3.6-27B}; \href{https://huggingface.co/Qwen/Qwen3-32B}{Qwen3-32B}; \href{https://huggingface.co/google/gemma-4-31B}{Gemma-4-31B}; \href{https://github.com/deepseek-ai/DeepSeek-R1}{DeepSeek-R1}; and \href{https://huggingface.co/JH976/Perovskite-R1}{Perovskite-R1}.

\endgroup
\end{landscape}

\clearpage
\suppfiguretocshort{S1}{Construction of the OpenPerov scientific corpus}
\suppfiguretocshort{S2}{Journal, publication period and article-type composition}
\suppfiguretocshort{S3}{Scientific topic profile of the OpenPerov corpus}
\suppfiguretocshort{S4}{Citation lineage and scientific specialization}
\suppfiguretocshort{S5}{Evidence-aware synthesis and claim calibration}
\suppfiguretocshort{S6}{Design and scientific scope of PSM-Bench}
\suppfiguretocshort{S7}{Empirical difficulty and discrimination in PSM-Bench}
\suppfiguretocshort{S8}{Criterion-level evaluation and robust aggregation}
\suppfiguretocshort{S9}{Closed-book prompt and information boundary}
\suppfiguretocshort{S10}{Human-expert validation of PSM-Bench scoring}
\suppfiguretocshort{S11}{Prompt robustness of closed-book evaluation}
\suppfiguretocshort{S12}{Conventional versus structured scientific adaptation}
\suppfiguretocshort{S13}{Contributions of DAPT and scientific training}
\suppfiguretocshort{S14}{Cross-scale adaptation and backbone selection}
\suppfiguretocshort{S15}{Model-resolved PSM-Bench performance}
\suppfiguretocshort{S16}{Item-level robustness of the OpenPerov Flash advantage}
\suppfiguretocshort{S17}{Scientific-evidence normalization for OpenPerov Pro}
\suppfiguretocshort{S18}{Retrieval and evidence access in OpenPerov Pro}
\suppfiguretocshort{S19}{Local evidence revision: gain--harm balance}
\suppfiguretocshort{S20}{Post-cutoff scientific reasoning under two information settings}
\suppfiguretocshort{S21}{Evolution of perovskite-stability research}
\suppfiguretocshort{S22}{Forward signals in perovskite-stability research}
\suppfiguretocshort{S23}{Transfer to the Perovskite-R1 benchmark}
\suppfiguretocshort{S24}{Photovoltaic performance of BCPCz--TMPS--M devices}
\phantomsection
\addtocontents{toc}{\protect\newpage\protect\contentscontinued}
\addcontentsline{toc}{section}{Supplementary Figures}

\suppfigurepage{S1}
{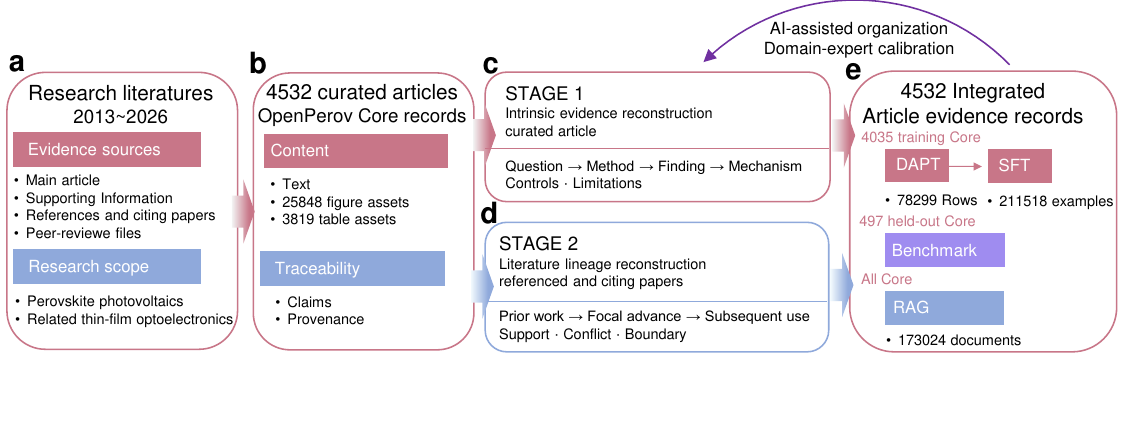}
{Construction of the OpenPerov article-centred scientific corpus}
{The corpus comprises 4,532 curated articles published in 2013--2026, with 25,848 figure and 3,819 table assets. Domain experts reconstructed and calibrated article evidence and literature lineage in 4,532 Integrated Article Evidence Records, with GPT-5.5 and GPT-5.6 Sol assisting with text processing. The 4,035 adaptation articles supplied 78,299 structured evidence records and 211,518 PDF-derived records for the conventional comparison. Both routes used the same source articles and backbone, followed their respective initial adaptation objectives, and then received the same scientific instruction and answer-style training. The structured route produced Flash. The remaining 497 articles were held out from adaptation, and training content that overlapped with their key scientific conclusions was removed. Pro retrieves from 173,024 documents covering all 4,532 articles. Adaptation data excluded benchmark answers (SM1--SM2).}

\suppfigurepage{S2}
{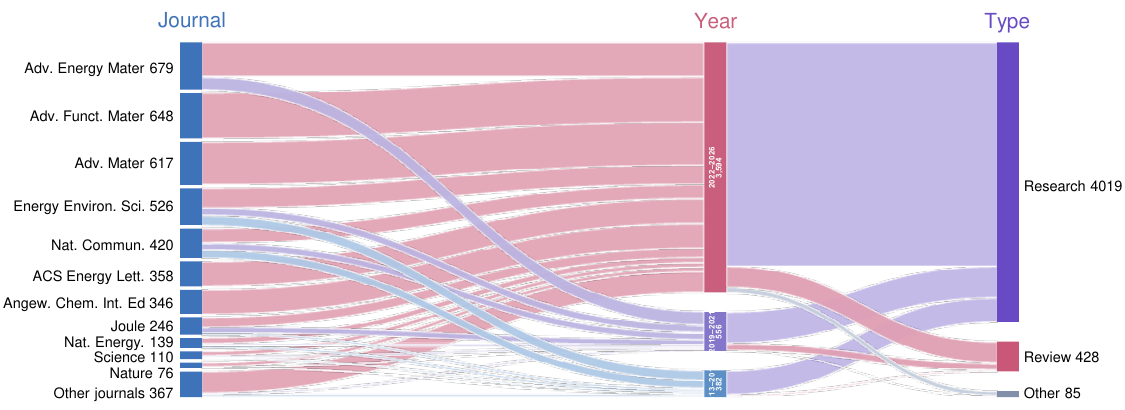}
{Journal, publication-period and article-type composition of the OpenPerov corpus}
{Alluvial diagram of 4,532 CORE articles. The journal column includes the ten most frequently represented journals and \textit{Nature}, with the remaining titles grouped as Other journals. Publication periods are 2013--2018 ($n=382$), 2019--2021 ($n=556$) and 2022--2026 ($n=3{,}594$). Harmonized article types comprise research records ($n=4{,}019$), reviews ($n=428$) and other types ($n=85$). Ribbon widths indicate article counts between adjacent categories.}

\suppfigurepage{S3}
{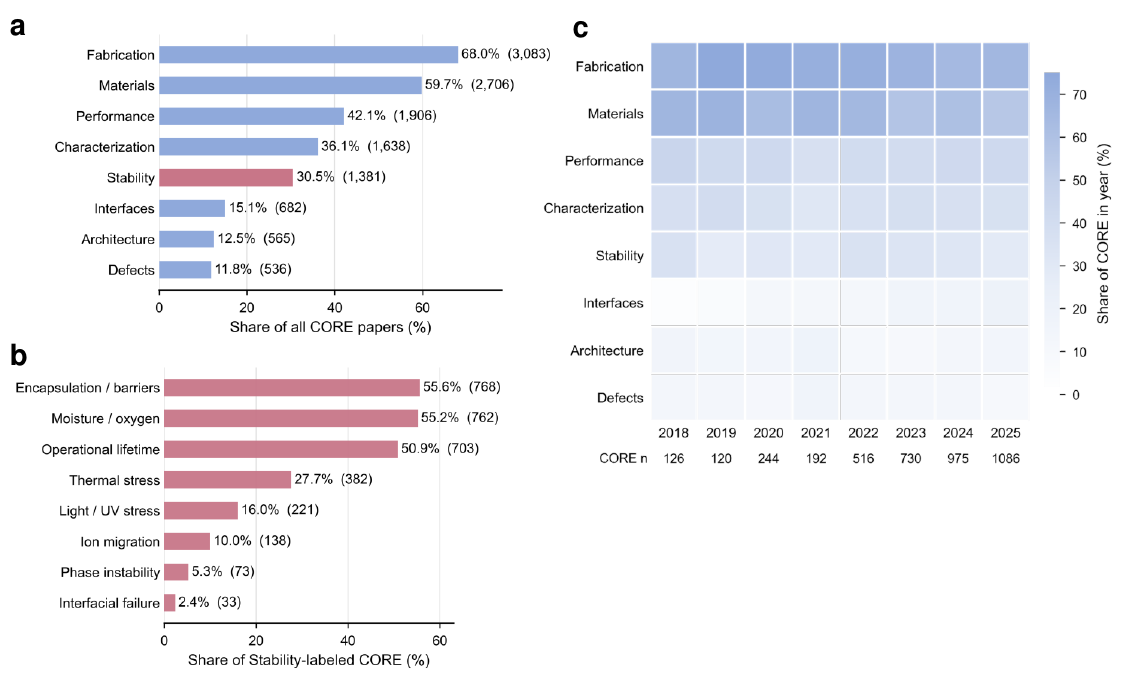}
{Scientific topic profile of the OpenPerov corpus}
{\textbf{a,} Prevalence of eight themes among 4,532 CORE papers, with counts in parentheses. \textbf{b,} Stability subtopics among 1,381 papers assigned to stability and degradation. \textbf{c,} Annual theme prevalence for 2018--2025, normalized to each year's CORE paper count (shown below the heat map). Themes were assigned from article evidence using predefined tag mappings and support scores (SM8). Papers may belong to multiple themes and stability subtopics.}

\suppfigurepage{S4}
{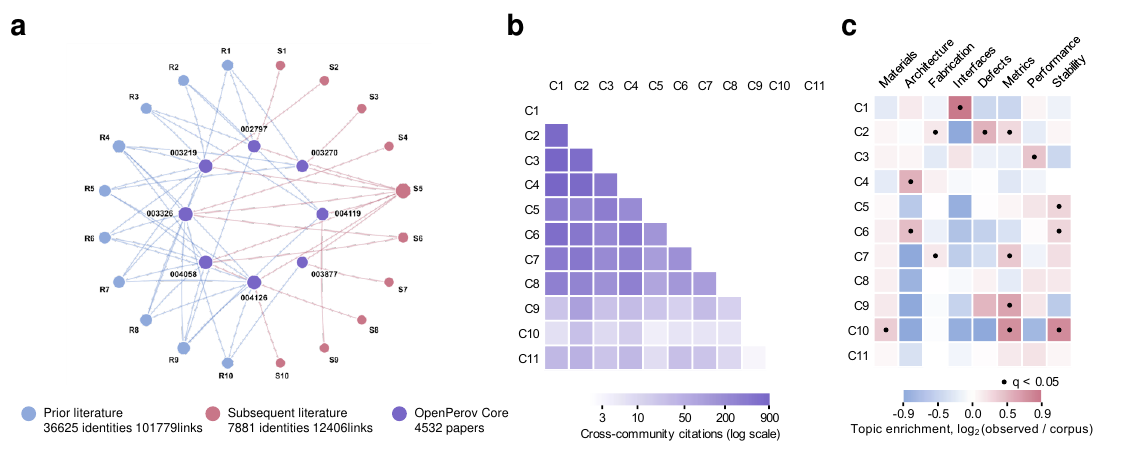}
{Citation lineage and scientific specialization of the OpenPerov corpus}
{\textbf{a,} Article-centred lineage spanning 4,532 focal articles, 36,625 distinct cited-paper identities linked by 101,779 reference relations, and 7,881 citing-paper identities linked by 12,406 relations. \textbf{b,} Citation communities within the focal-article network. The largest weakly connected component contains 3,758 papers in 11 communities; cells show cross-community citation counts on a logarithmic colour scale. \textbf{c,} Theme enrichment in these communities. Black dots mark positive enrichment after Benjamini--Hochberg correction ($q<0.05$). The communities differ in scientific emphasis while sharing themes.}

\suppfigurepage{S5}
{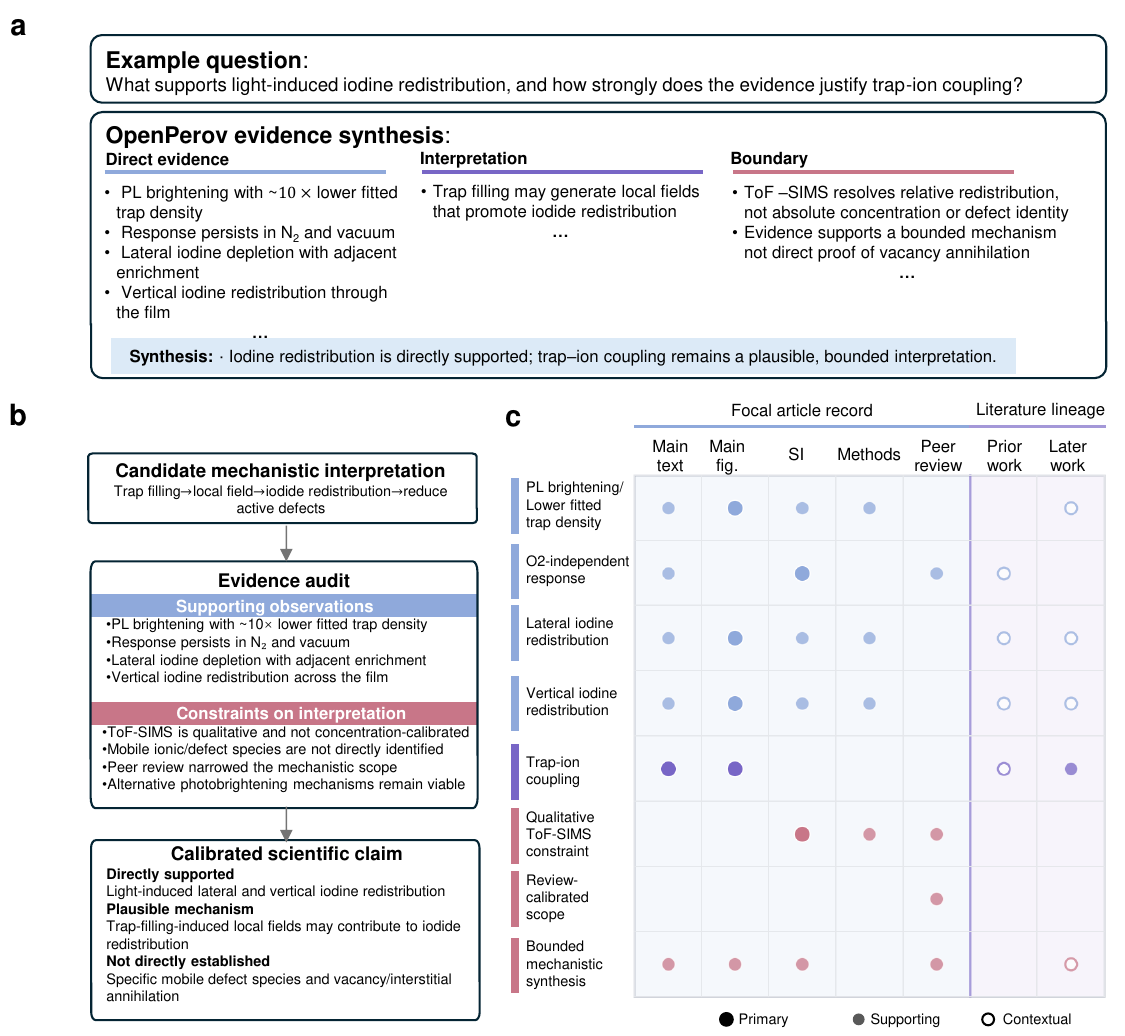}
{Evidence synthesis and claim calibration in a representative study}
{\textbf{a,} Research question and evidence-synthesis excerpts for light-induced iodine redistribution in MAPbI$_3$ films, linking observations to mechanistic interpretation. \textbf{b,} Evidence supporting the proposed mechanism and defining its scope. Photoluminescence brightening, reduced fitted trap density, an oxygen-independent response, and lateral and vertical iodine redistribution support the interpretation. The remaining questions are defined by the qualitative nature of ToF-SIMS, unresolved defect identities, refinement of claims during peer review and alternative mechanisms. \textbf{c,} Provenance across the focal article and its literature lineage. Colour denotes claim class; marker styles distinguish primary, supporting and contextual links.}

\suppfigurepage{S6}
{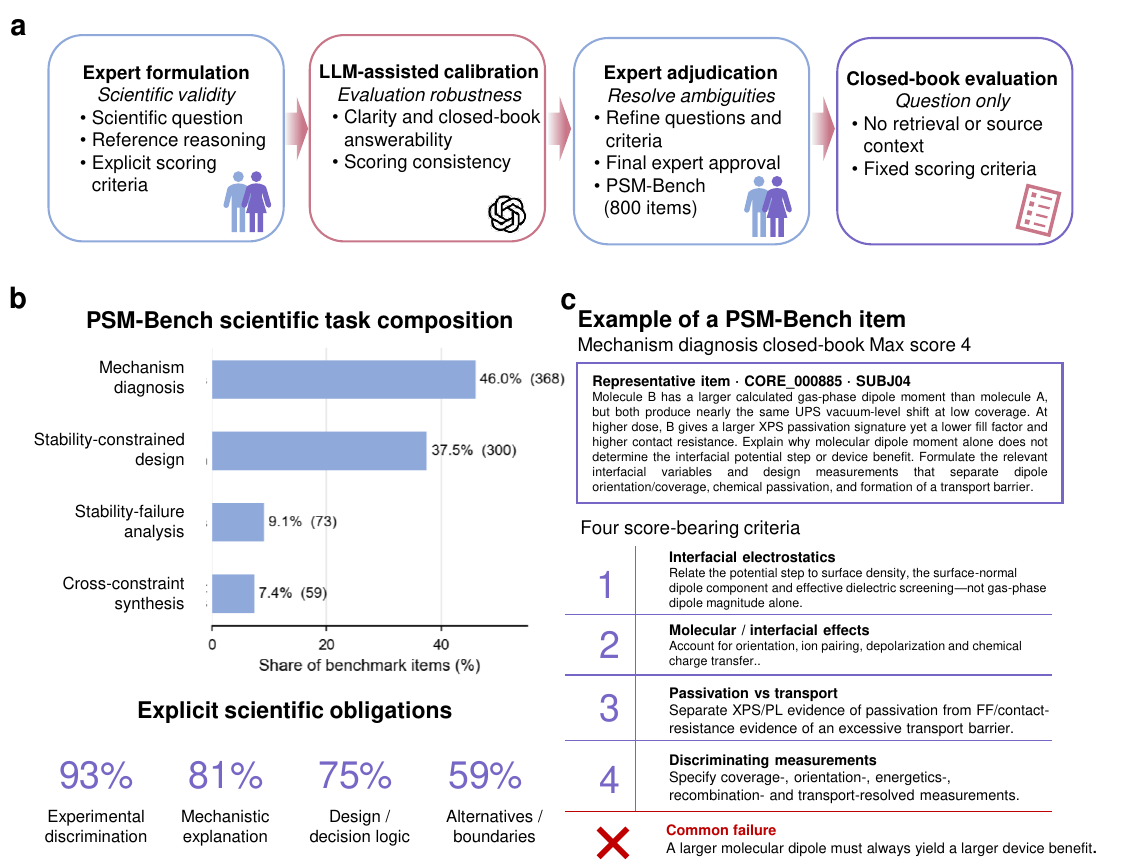}
{Design and scientific scope of PSM-Bench}
{\textbf{a,} Expert formulation, LLM-assisted calibration and expert adjudication before the 800-item benchmark was frozen. GPT-5.5 xhigh and GPT-5.6 Sol Pro assisted with checks of clarity, answerability and scoring consistency. \textbf{b,} Task composition: mechanism diagnosis ($n=368$), stability-constrained design ($n=300$), stability-failure analysis ($n=73$) and cross-constraint synthesis ($n=59$), alongside the prevalence of explicit reasoning requirements. \textbf{c,} Example mechanism-diagnosis question and its four scientific scoring criteria.}

\suppfigurepage{S7}
{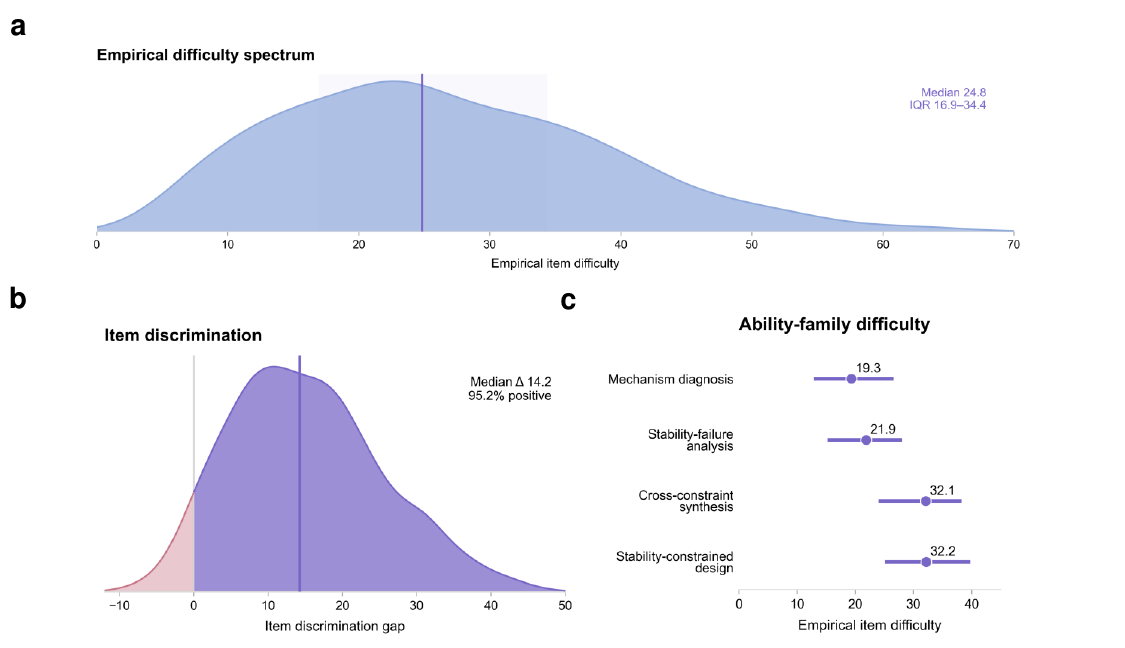}
{Empirical difficulty and discrimination in PSM-Bench}
{\textbf{a,} Item difficulty calculated from non-OpenPerov closed-book scores across 800 questions (median, 24.8; interquartile range, 16.9--34.4). \textbf{b,} Discrimination between stronger and weaker systems in the frozen 13-system reference panel described in SM4. The median gap is 14.2 points, with positive discrimination for 95.2\% of items. \textbf{c,} Difficulty by task family. Points show medians and horizontal bars show interquartile ranges. Median difficulty is higher for cross-constraint synthesis and stability-constrained design than for mechanism diagnosis and stability-failure analysis (SM4).}

\suppfigurepage{S8}
{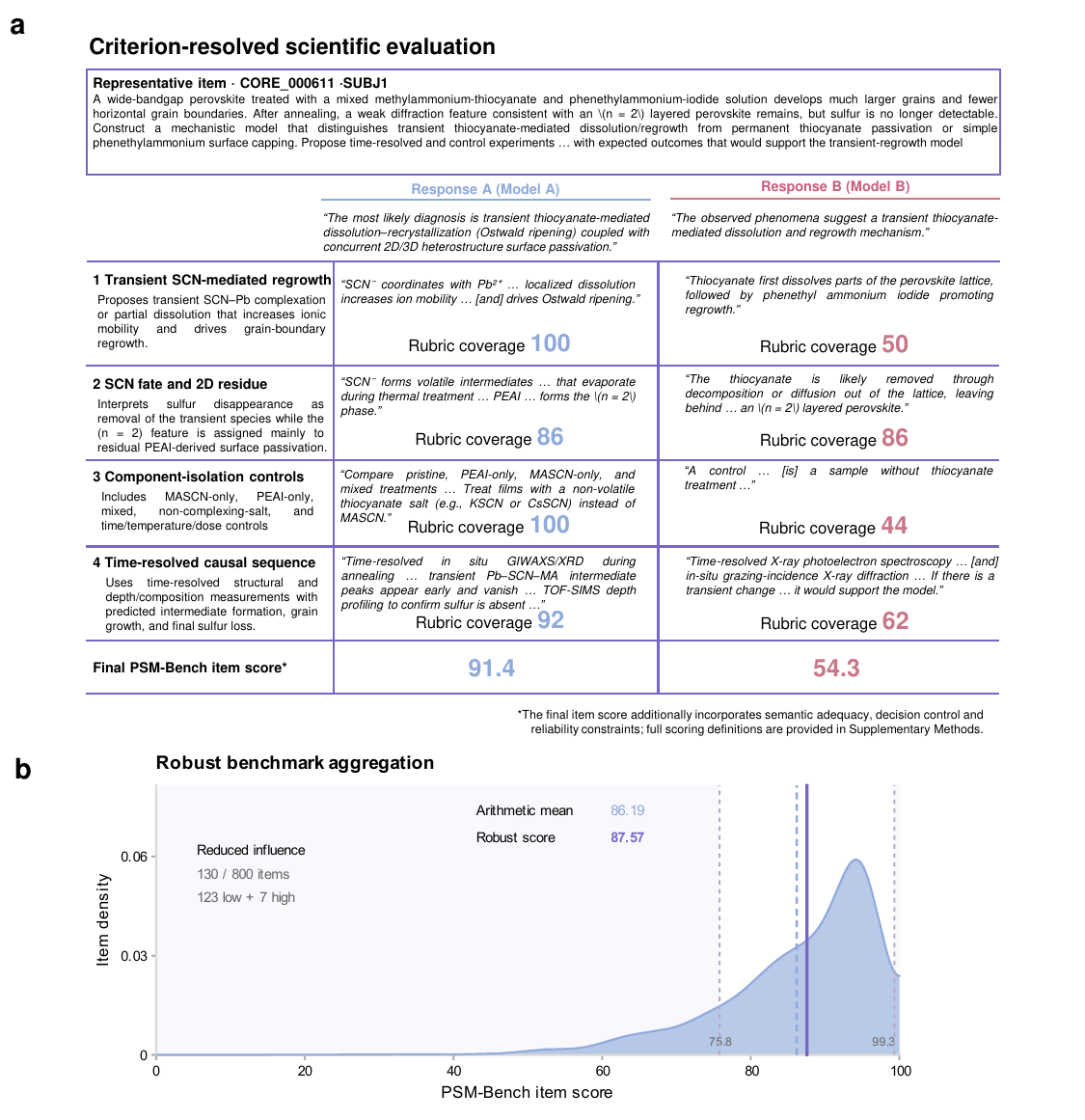}
{Criterion-level evaluation and robust aggregation of PSM-Bench scores}
{\textbf{a,} Two frozen responses to the same question assessed against four criteria. Both identify transient thiocyanate-mediated dissolution and regrowth but differ in mechanistic detail, component-isolation controls and time-resolved predictions. \textbf{b,} Flash scores across 800 items. Huber aggregation downweights 130 items (123 in the lower tail and 7 in the upper tail) while retaining all responses, yielding 87.57 compared with an arithmetic mean of 86.19 (SM5).}

\suppfigurepage{S9}
{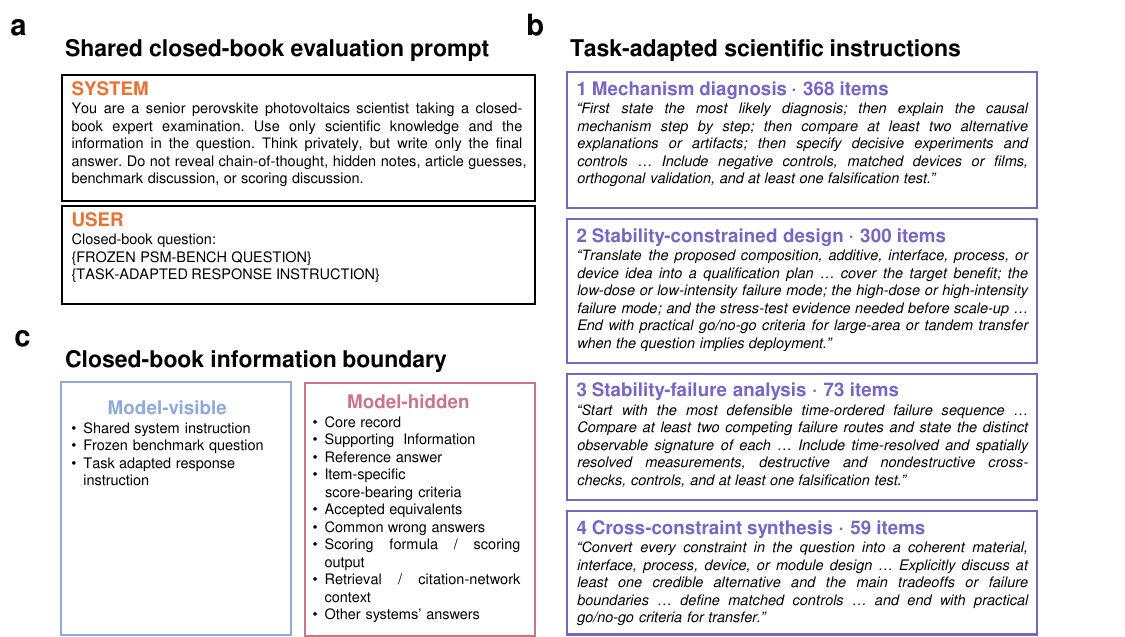}
{Closed-book prompts and information access in PSM-Bench}
{\textbf{a,} Shared prompt comprising the system instruction, frozen question and task-adapted instruction. \textbf{b,} Instructions for the four task families. \textbf{c,} Model-visible and withheld information. Prompts specified the requested scientific reasoning; source evidence, reference answers, item-specific criteria and retrieval context were withheld (SM4).}

\suppfigurepage{S10}
{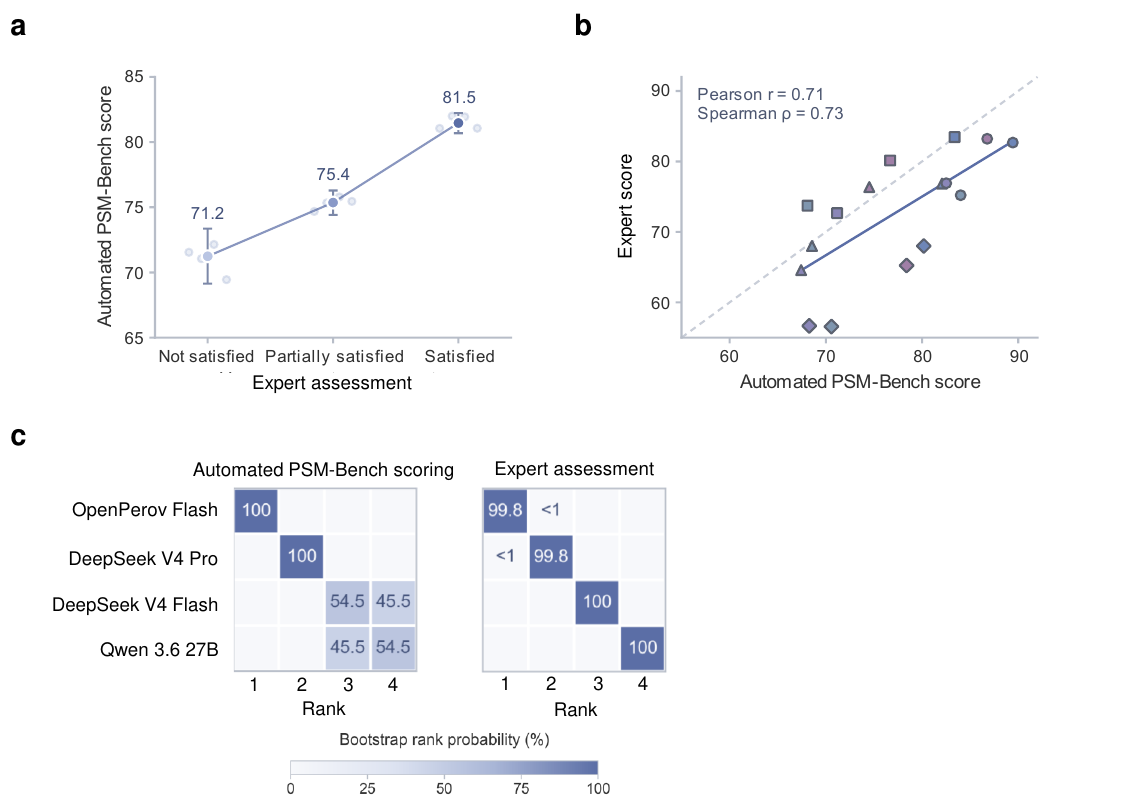}
{Human-expert validation of PSM-Bench scoring}
{\textbf{a,} Automated scores and expert-assessed criterion fulfilment. Large points show means across four criteria, pale points show individual-criterion means, and error bars show 95\% bootstrap confidence intervals. \textbf{b,} Agreement across system-by-family means (Pearson $r=0.71$; Spearman $\rho=0.73$). Dashed and solid lines show equality and the fitted relationship, respectively. \textbf{c,} Bootstrap rank frequencies: Flash ranked first in 100\% of automated-score and 99.8\% of expert-score resamples. The expert assessment was double-blind, with system identities and automated scores withheld; expert ratings were analysed separately (SM5).}

\suppfigurepage{S11}
{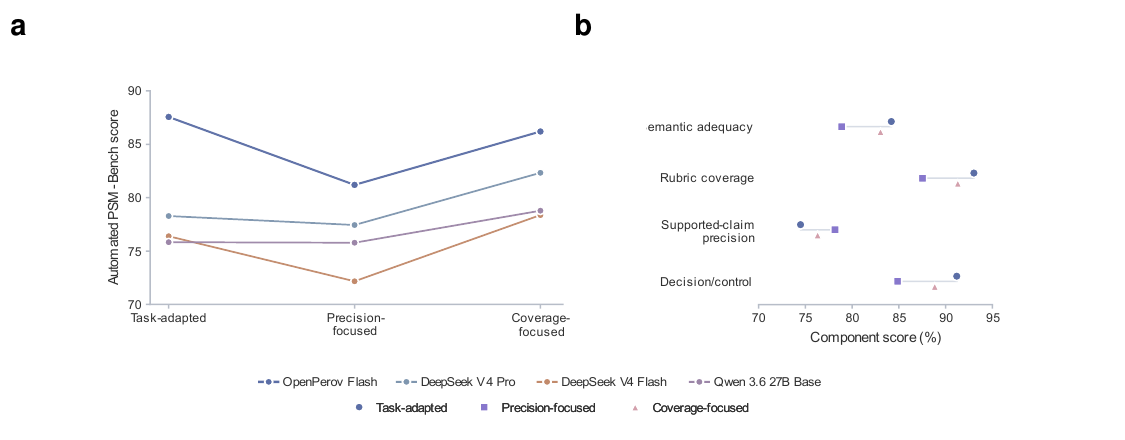}
{Prompt robustness of closed-book PSM-Bench evaluation}
{\textbf{a,} Scores for four systems under task-adapted, precision-focused and coverage-focused prompts. Flash ranked first and DeepSeek V4 Pro second in all conditions. \textbf{b,} Flash scoring components. The task-adapted prompt favoured semantic adequacy, rubric coverage and decision/control quality; the precision-focused prompt favoured supported-claim precision. Scores are displayed on a 0--100 scale; supported-claim precision is displayed as $100P$.}

\suppfigurepage{S12}
{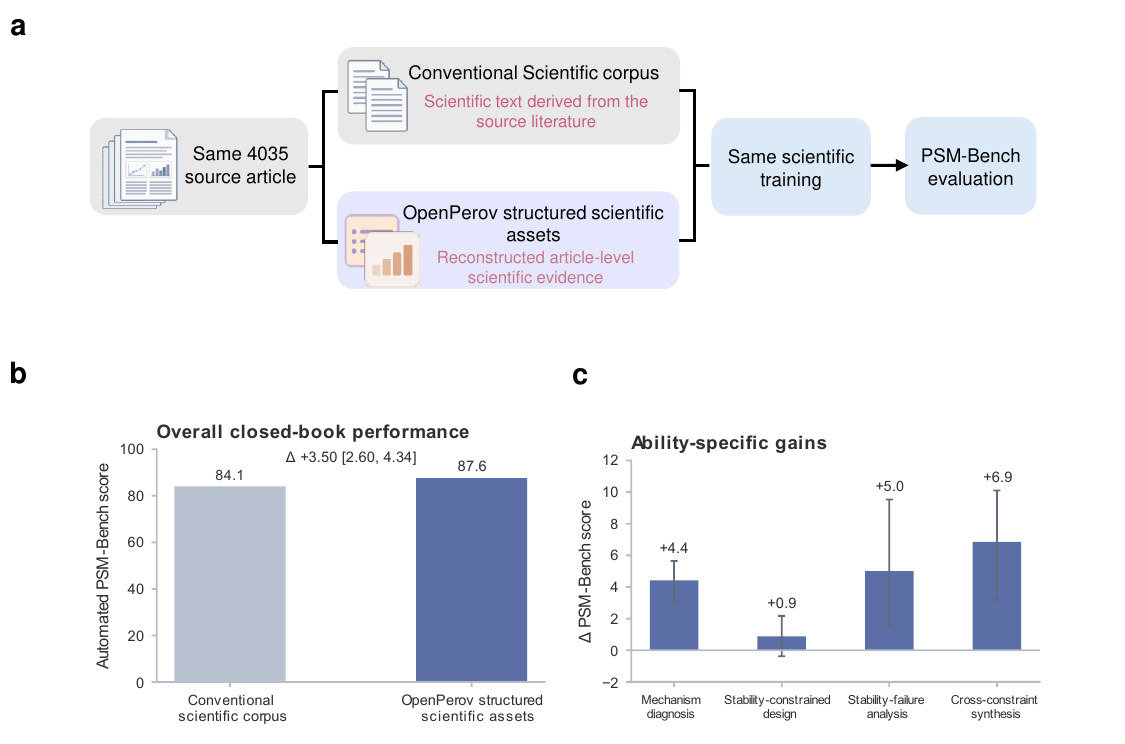}
{Conventional and structured scientific adaptation from the same source articles}
{\textbf{a,} Conventional PDF-derived text and reconstructed evidence from the same 4,035 articles were used to adapt the same backbone. Each route used its own initial adaptation objective, followed by the same scientific instruction and answer-style training (SM3). \textbf{b,} Final closed-book scores were 84.07 and 87.57, respectively, giving a paired gain of 3.50 points (95\% confidence interval, 2.60--4.34). \textbf{c,} Gains across all four task families, largest for cross-constraint synthesis, stability-failure analysis and mechanism diagnosis. Error bars show paired 95\% item-bootstrap confidence intervals.}

\suppfigurepage{S13}
{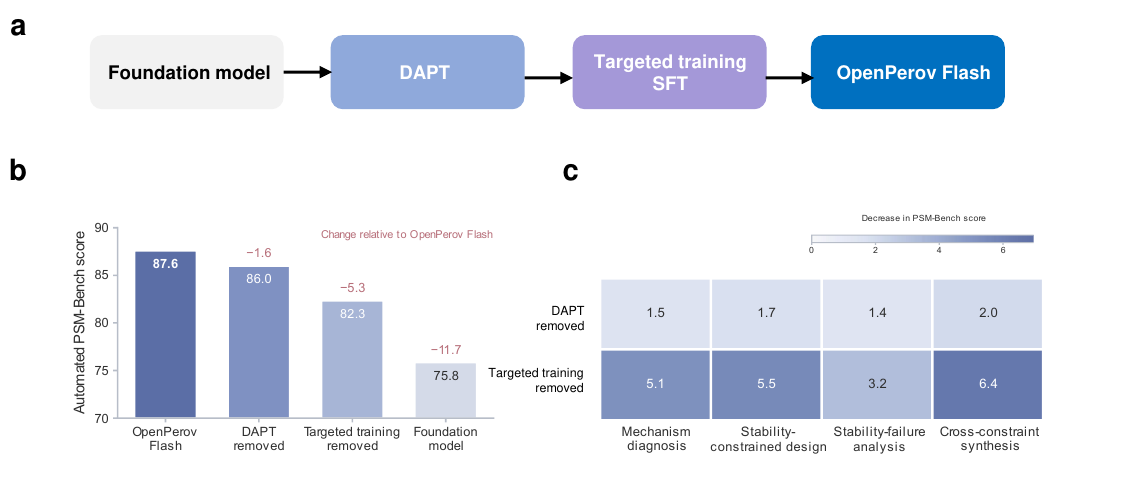}
{Contributions of DAPT and targeted scientific training to OpenPerov Flash}
{\textbf{a,} Training sequence from the foundation model through DAPT and targeted scientific training. \textbf{b,} Overall scores for the full training sequence and component ablations. \textbf{c,} Score decreases by task family after each training component was removed. DAPT provided consistent gains, while targeted training had a larger effect, especially for stability-constrained design and cross-constraint synthesis. Here, targeted training comprises both scientific-instruction and answer-style training; ``Targeted training removed'' therefore denotes the DAPT-only checkpoint (SM3).}

\suppfigurepage{S14}
{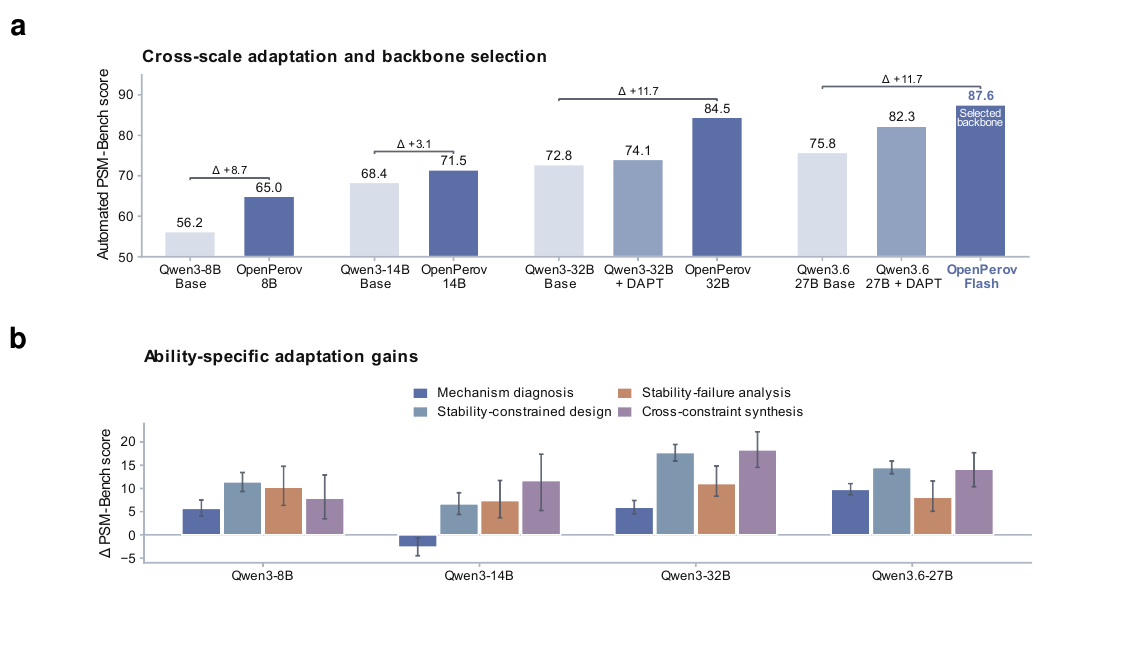}
{Cross-scale adaptation and backbone selection for OpenPerov Flash}
{\textbf{a,} Closed-book scores across backbones and training states. Adaptation improved every scale tested, and Qwen3.6-27B and Qwen3-32B showed nearly identical gains. Starting from a higher score, Qwen3.6-27B reached the highest score after training, supporting its selection for Flash. \textbf{b,} Gains across the four task families. Error bars show 95\% paired item-bootstrap intervals from 10,000 resamples. Scores are Huber location estimates with $c=1.345$.}

\suppfigurepage{S15}
{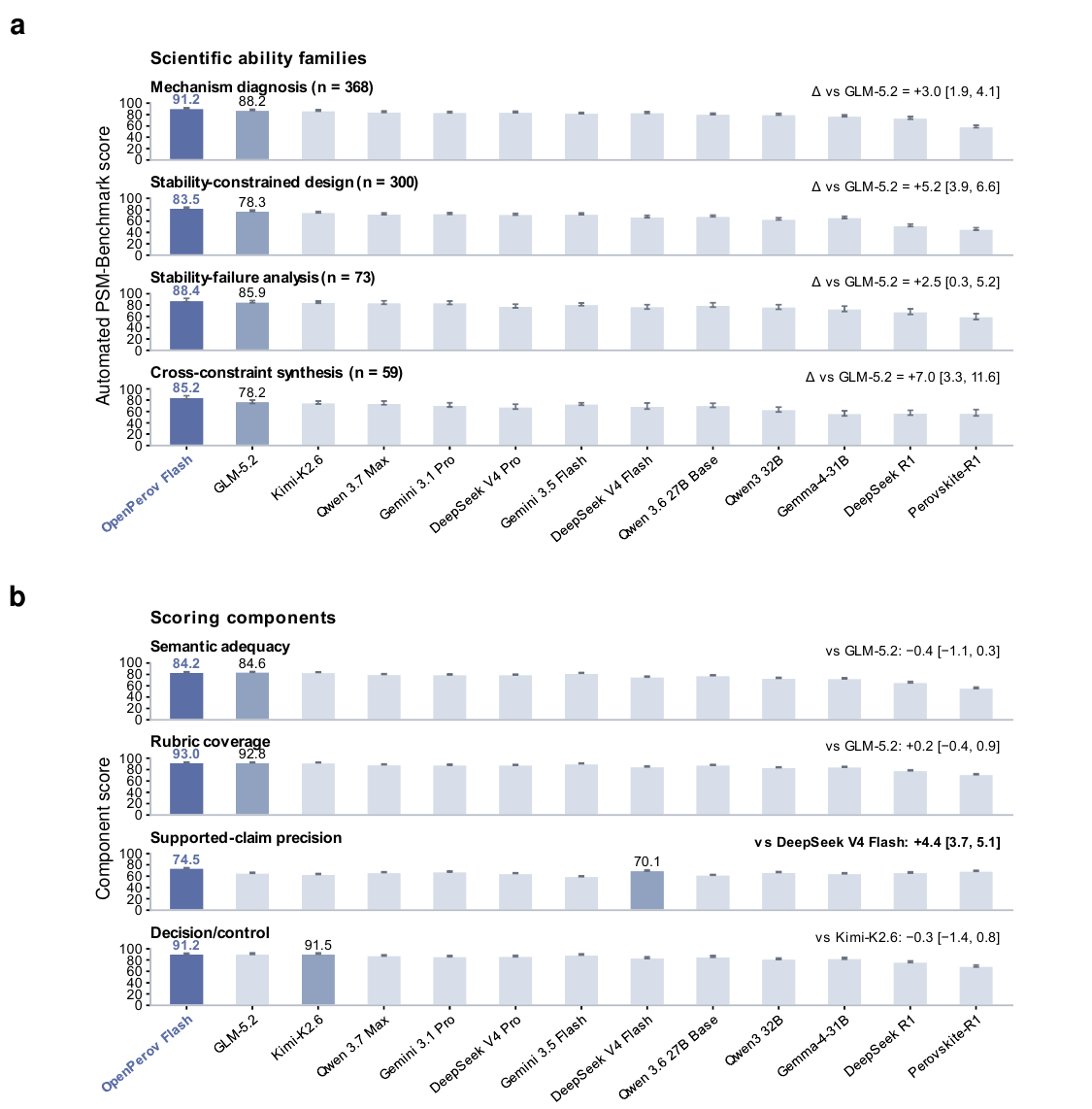}
{Model-level PSM-Bench performance}
{\textbf{a,} Closed-book scores by task family, with 95\% item-bootstrap confidence intervals. Flash scored highest in all four families, with its largest margins over GLM-5.2 in stability-constrained design and cross-constraint synthesis. \textbf{b,} Semantic adequacy, rubric coverage, supported-claim precision and decision/control scores. Flash was among the leading systems on each component, with its clearest advantage in supported-claim precision. Model order is consistent across panels.}

\suppfigurepage{S16}
{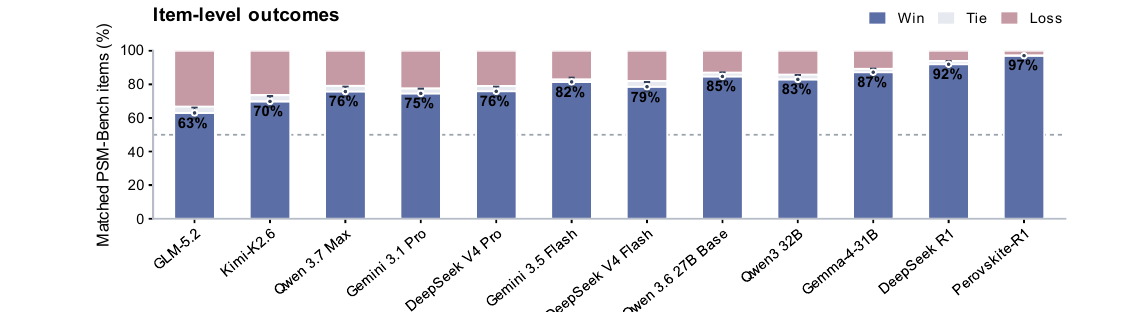}
{Item-level comparisons of OpenPerov Flash with closed-book systems}
{Fractions of the 800 questions on which Flash scored higher than, the same as or lower than each comparator. Points and error bars show win fractions and 95\% paired item-bootstrap confidence intervals. Flash scored higher on a majority of questions against every comparator, including 63\% against GLM-5.2.}

\suppfigurepage{S17}
{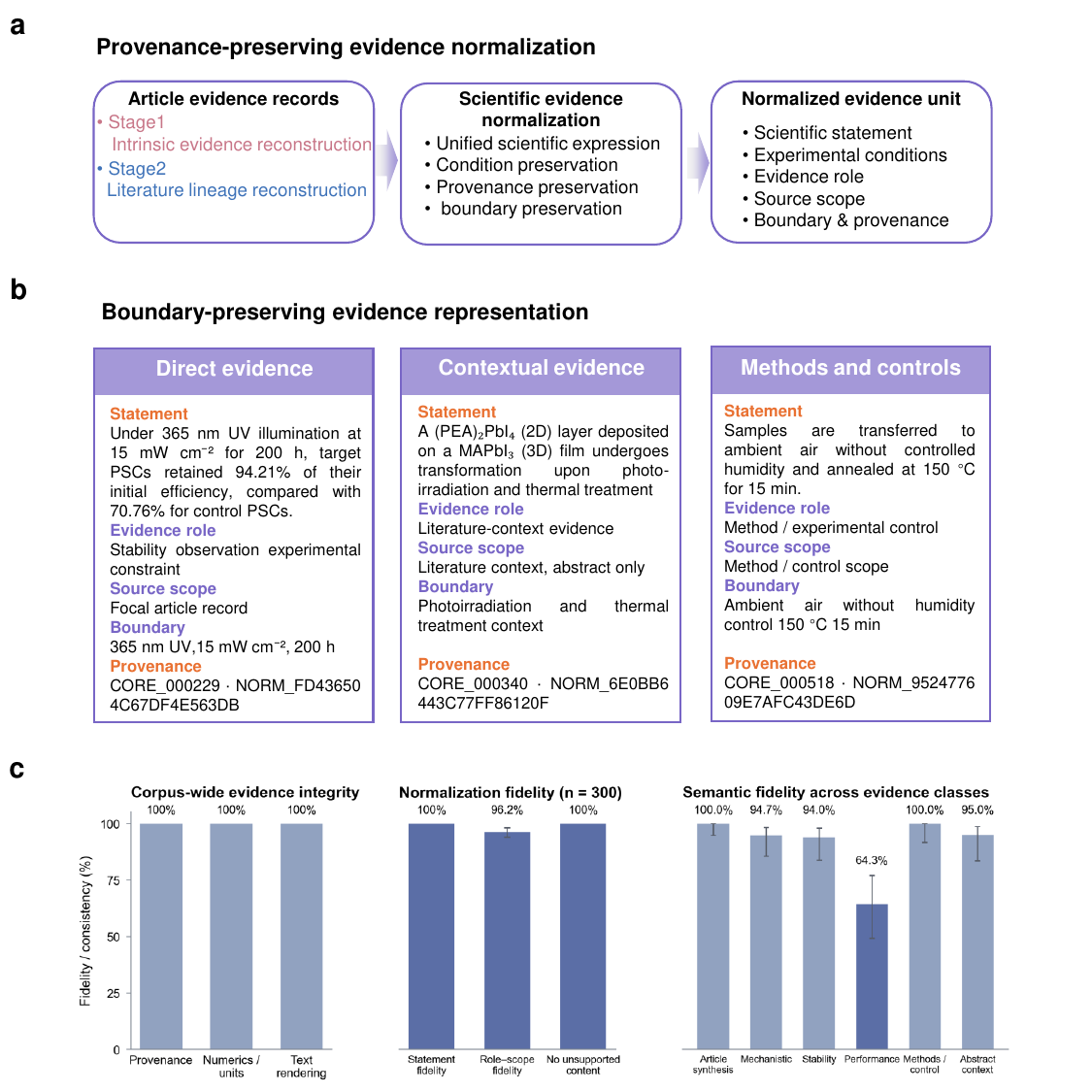}
{Source-preserving normalization of scientific evidence for OpenPerov Pro}
{\textbf{a,} Conversion of article evidence and literature lineage into a common representation that retains conditions, source scope and provenance. \textbf{b,} Examples of direct evidence, contextual evidence, and methods or controls. \textbf{c,} Corpus-wide integrity checks and a stratified expert audit. Corpus-wide checks confirmed preservation of provenance, numerical information and rendered text. Scientific content was preserved in all 300 reviewed blocks (statement fidelity, 300/300). The right panel reports agreement between block-level evidence roles and source strata (role--scope consistency) by evidence class, including 27/42 for performance evidence.}

\suppfigurepage{S18}
{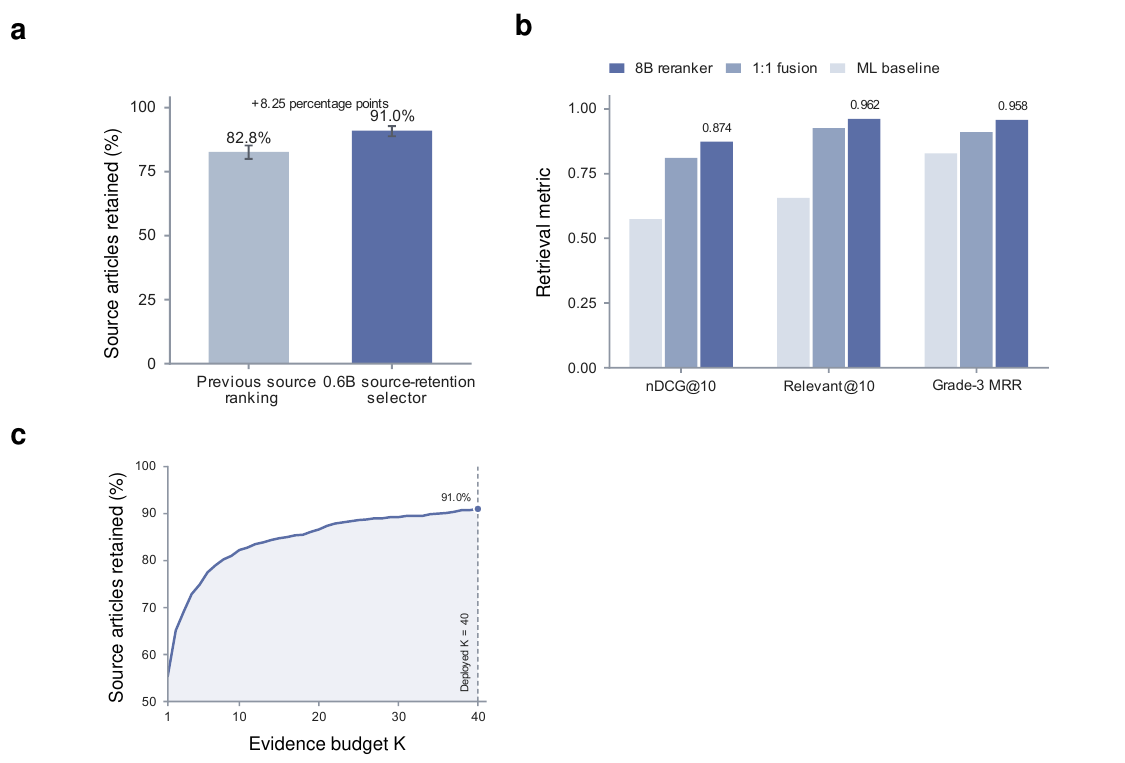}
{Retrieval and evidence access in OpenPerov Pro}
{\textbf{a,} The 0.6B source-retention selector retained the source article in the Top 40 for 728/800 items (91.00\%), compared with 662/800 (82.75\%) under the preceding ranking, an 8.25-percentage-point gain. \textbf{b,} Performance of the 8B scientific-relevance reranker, fixed 1:1 fusion and machine-learning baseline on identical 40-paper pools with expert-corrected judgments. The reranker achieved nDCG@10 of 0.874, Relevant@10 of 0.962 and Grade-3 MRR of 0.958. \textbf{c,} Source retention as a function of budget $K$, with $K=40$ used in Pro (SM6).}

\suppfigurepage{S19}
{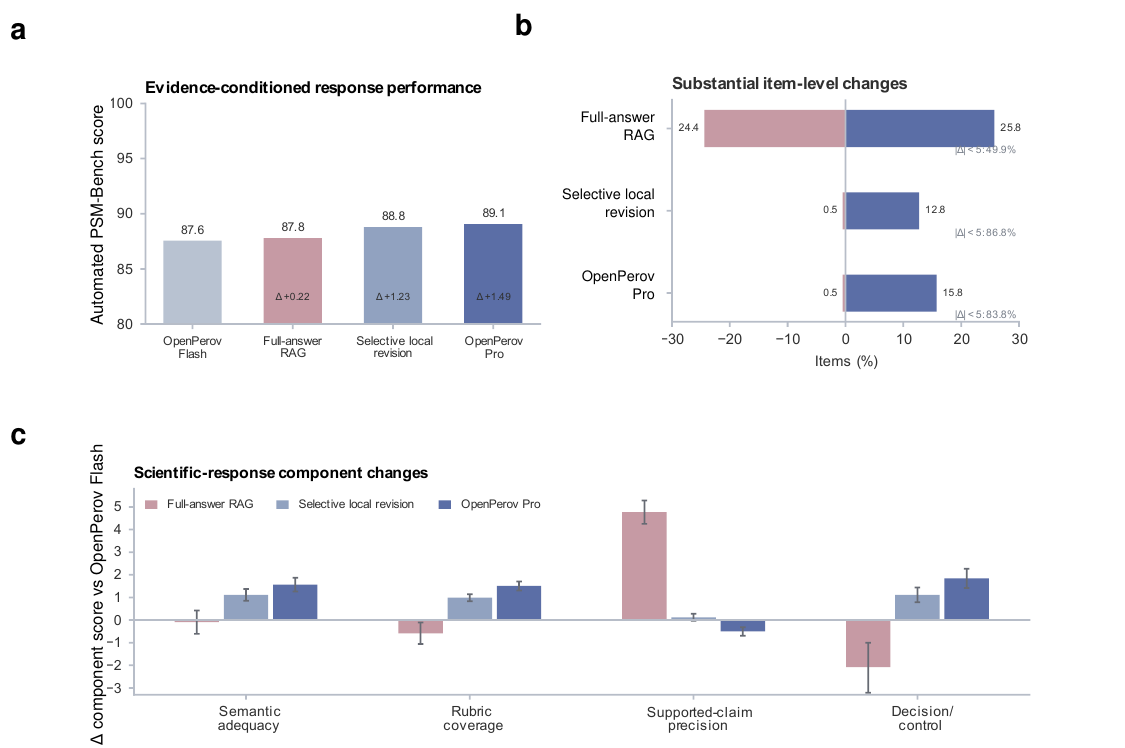}
{Local evidence revision improves the gain--harm balance}
{\textbf{a,} Flash, full-answer RAG, selective local revision and Pro on 800 matched items. Relative to Flash, robust-score gains were $+0.22$, $+1.23$ and $+1.49$ points for RAG, selective revision and Pro, respectively. \textbf{b,} Substantial changes at the $\pm5$-point threshold: gains/losses of 25.75\%/24.38\%, 12.75\%/0.50\% and 15.75\%/0.50\% for the same three routes. \textbf{c,} Changes in scoring components relative to Flash, with paired 95\% item-bootstrap confidence intervals. Selective revision denotes the critic-selected alternative to Pro (SM6).}

\suppfigurepage{S20}
{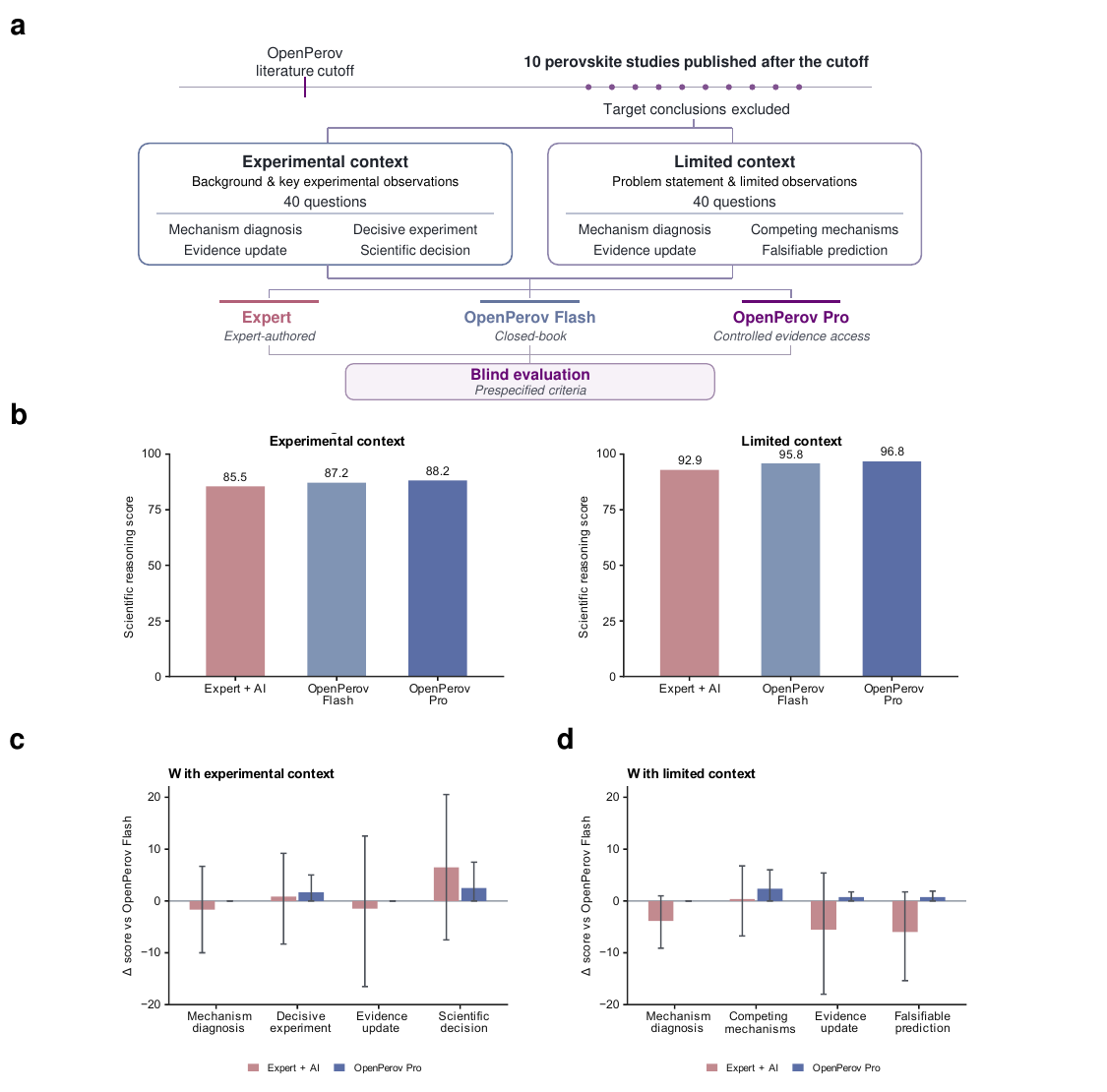}
{Post-cutoff scientific reasoning under two information settings}
{\textbf{a,} Ten post-cutoff studies supplied 40 questions with Experimental context and 40 with Limited context; target conclusions were withheld. Experts could consult online resources available up to OpenPerov's literature cutoff. Independent volunteer perovskite experts finalized scores after evaluating anonymized responses and discussing their assessments (SM7). \textbf{b,} Scores in the two settings were 87.2/95.8 for Flash, 85.5/92.9 for expert-authored answers and 88.2/96.8 for Pro. The plotted label \emph{Expert + AI} denotes expert-authored answers (Methods). Each setting used its own rubric. \textbf{c,d,} Task-level differences from Flash. Pro gains were concentrated in the decisive-experiment and scientific-decision tasks with Experimental context (\textbf{c}), and the competing-mechanisms task with Limited context (\textbf{d}), while preserving mechanism-diagnosis performance. Error bars show paired 95\% study-bootstrap confidence intervals.}

\suppfigurepage{S21}
{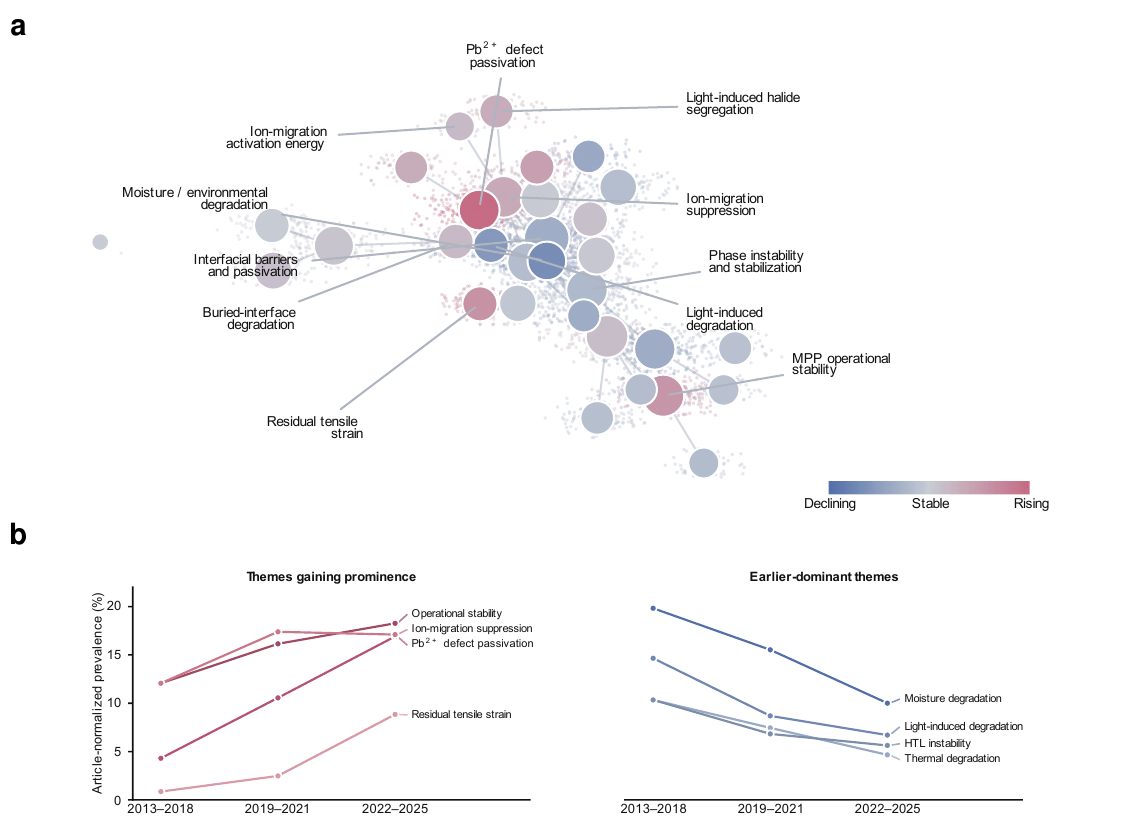}
{Evolution of perovskite-stability research}
{\textbf{a,} Evidence-derived research landscape. Large nodes denote expert-reviewed themes; node size indicates article coverage, and colour indicates whether article-normalized prevalence declined or increased. Small points denote canonical concepts, and grey edges show network relationships. \textbf{b,} Trajectories in 2013--2018, 2019--2021 and 2022--2025. Pb$^{2+}$ passivation, residual tensile strain, operational stability and ion-migration suppression gained prominence. By contrast, moisture, light-induced and thermal degradation, together with hole-transport-layer (HTL) instability, declined in relative prevalence.}

\suppfigurepage{S22}
{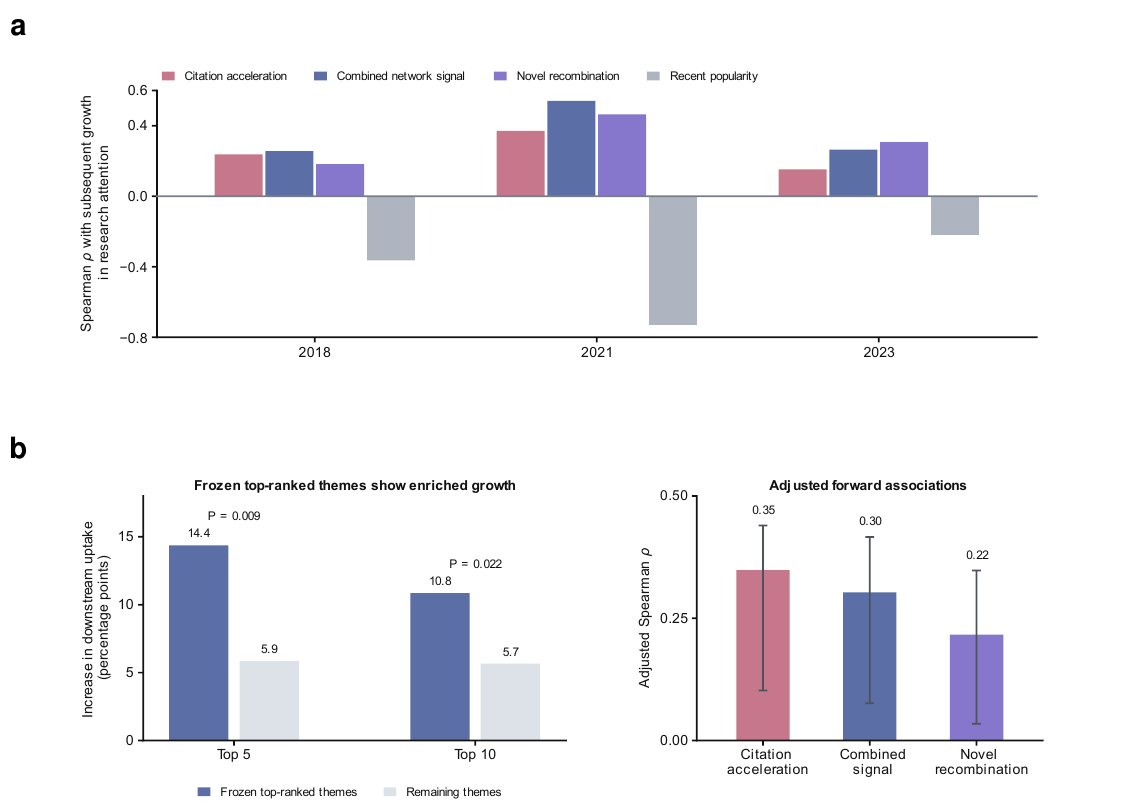}
{Signals of subsequent growth in perovskite-stability research attention}
{\textbf{a,} Historical analyses at the 2018, 2021 and 2023 cutoffs. Citation acceleration, the combined signal and novel recombination correlated positively with subsequent changes in downstream uptake; recent popularity correlated negatively. \textbf{b,} The top-five and top-ten themes in the frozen 2025 ranking gained more uptake in the partial-2026 literature than the remaining themes (left). After adjustment for theme size and recent popularity, citation acceleration showed the strongest positive association, followed by the combined signal and novel recombination (right). Error bars show 95\% paper-bootstrap confidence intervals (SM8).}

\suppfigurepage{S23}
{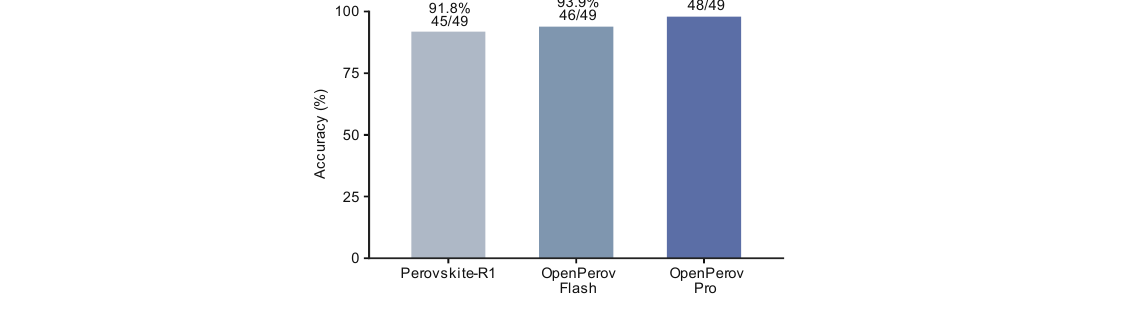}
{Transfer to the independent Perovskite-R1 benchmark}
{Exact-match accuracy on the 49 multiple-choice questions released with Perovskite-R1~\citep{wang2026perovskiteR1}. Perovskite-R1 answered 45/49 correctly (91.8\%), Flash 46/49 (93.9\%) and Pro 48/49 (98.0\%). Perovskite-R1 predictions came from the authors' release; OpenPerov predictions came from local runs (SM7).}

\suppfigurepage{S24}
{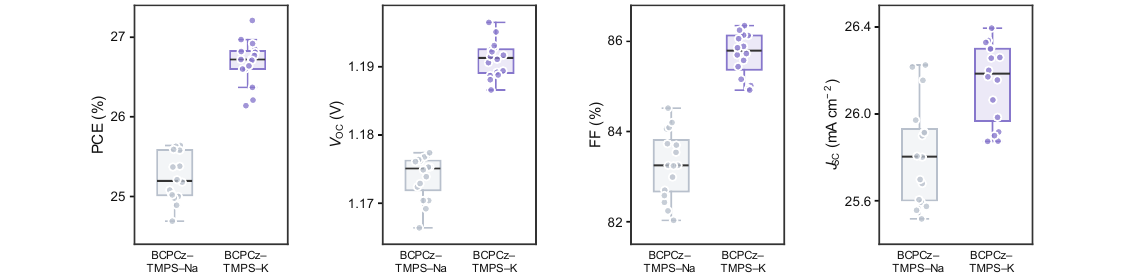}
{Photovoltaic performance of BCPCz--TMPS--M devices}
{Power conversion efficiency (PCE), open-circuit voltage ($V_\mathrm{OC}$), fill factor (FF) and short-circuit current density ($J_\mathrm{SC}$) for BCPCz--TMPS--Na and BCPCz--TMPS--K devices ($n=16$ independent devices per group). Boxes show medians and interquartile ranges; whiskers reach the most extreme values within 1.5 times that range. Points represent individual devices. Replacing Na with K increased mean PCE from 25.24\% to 26.69\%, $V_\mathrm{OC}$ from 1.174 to 1.191 V, FF from 83.28\% to 85.71\%, and $J_\mathrm{SC}$ from 25.82 to 26.15 mA cm$^{-2}$.}

\end{document}